\documentclass[letterpaper, 10 pt, journal, twoside]{ieeetran}  

\IEEEoverridecommandlockouts                              
\makeatletter
\def\bstctlcite{\@ifnextchar[{\@bstctlcite}{\@bstctlcite[@auxout]}}
\def\@bstctlcite[#1]#2{\@bsphack
  \@for\@citeb:=#2\do{%
    \edef\@citeb{\expandafter\@firstofone\@citeb}%
    \if@filesw\immediate\write\csname #1\endcsname{\string\citation{\@citeb}}\fi}%
  \@esphack}
\makeatother

\usepackage[vlined, ruled, linesnumbered, commentsnumbered]{algorithm2e}
\usepackage[table]{xcolor}
\usepackage{amsmath}
\usepackage{bm}
\usepackage{cite}
\usepackage{color}
\usepackage{xcolor}
\definecolor{chred}{rgb}{0.8,0,0}
\definecolor{chgrey}{rgb}{0.5,0.5,0.5}
\usepackage{graphicx}
\usepackage{subcaption}
\usepackage{dblfloatfix}
\usepackage{eqlist}
\usepackage{txfonts}
\usepackage{url}
\usepackage{footmisc}
\usepackage{booktabs}
\usepackage{balance}

\usepackage{multirow}
\usepackage[para,online,flushleft]{threeparttable}
\usepackage[final]{pdfpages}
\usepackage{flushend}

\begin{document}
\IEEEoverridecommandlockouts

\title{\LARGE \bf A Mechanical Screwing Tool for 2-Finger Parallel Grippers\\ -- Design, Optimization, and Manipulation Policies}

\author{Zhengtao Hu$^{1}$, Weiwei Wan$^{1*}$, Keisuke Koyama$^{1}$, and Kensuke Harada$^{1,2}$
\thanks{$^{1}$ Graduate School of Engineering Science, Osaka University, Toyonaka, Osaka, Japan. $^{2}$ National Inst. of AIST}
\thanks{$^{*}$ Contact: Weiwei Wan, {\tt\small wan@sys.es.osaka-u.ac.jp}}%
}

\maketitle

\begin{abstract}
This paper develops a mechanical tool as well as its manipulation policies for 2-finger parallel robotic grippers. It primarily focuses on a mechanism that converts the gripping motion of 2-finger parallel grippers into a continuous rotation to realize tasks like fastening screws. The essential structure of the tool comprises a Scissor-Like Element (SLE) mechanism and a double-ratchet mechanism. They together convert repeated linear motion into continuous rotating motion. At the joints of the SLE mechanism, elastic elements are attached to provide resisting force for holding the tool as well as for producing torque output when a gripper releases the tool. The tool is entirely mechanical, allowing robots to use the tool without any peripherals and power supply. The paper presents the details of the tool design, optimizes its dimensions and effective stroke lengths, and studies the contacts and forces to achieve stable grasping and screwing. Besides the design, the paper develops manipulation policies for the tool. The policies include visual recognition, picking-up and manipulation, and exchanging tooltips. The developed tool produces clockwise rotation at the front end and counter-clockwise rotation at the back end. Various tooltips can be installed at both two ends. Robots may employ the developed manipulation policies to exchange the tooltips and rotating directions following the needs of specific fastening or loosening tasks. Robots can also reorient the tool using pick-and-place or handover, and move the tool to work poses using the policies. The designed tool, together with the developed manipulation policies, are analyzed and verified in several real-world applications. The tool is small, cordless, convenient, and has good robustness and adaptability.
\end{abstract}

\begin{IEEEkeywords}
    Grippers and Other End-Effectors, Grasping, Manipulation Planning 
\end{IEEEkeywords}

\section{INTRODUCTION}

\IEEEPARstart{I}{n} the past, people have been developing both exchangeable robotic end-effectors, e.g., parallel grippers and suction cups, and dexterous robotic end-effectors, e.g., anthropomorphic hands, to meet the multi-purpose requirements in both industry and academia. Unfortunately, none of these end-effectors are general enough to meet all needs. The exchangeable end-effectors require extra tool-changing mechanisms and power cables or vacuum supply tubes to be firmly connected to the end of a robot arm. The exchanging process is time-consuming and fragile. On the other hand, dexterous hands are complicated, expensive, and difficult to control. Even if the control problem is solved, the generality of a dexterous hand is still unclear as they assimilate a human hand, and may not be more flexible than a human hand.
\begin{figure}[!t]
    \begin{center}
    \includegraphics[width=0.97\linewidth]{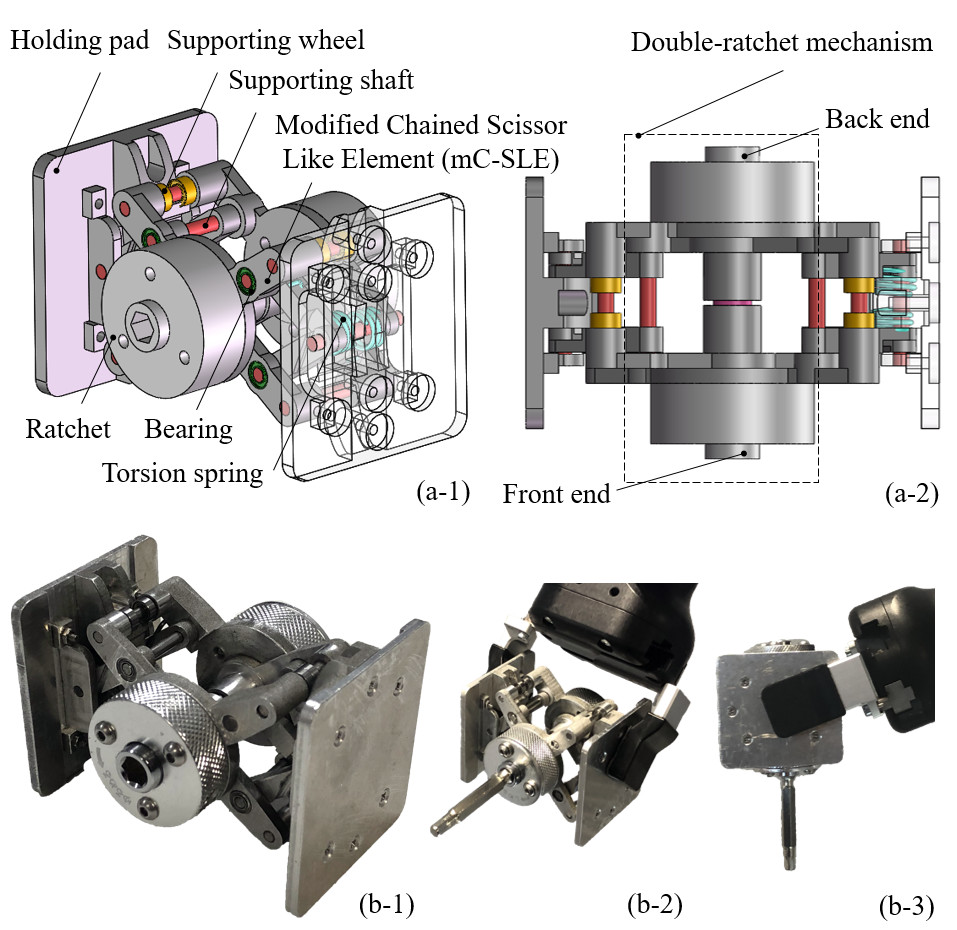}
    \caption{The proposed mechanical screwing tool. It employs a Scissor-Like Element (SLE) mechanism and a double-ratchet mechanism to convert parallel gripping motion to continuous rotating motion. (a-1,2) The CAD models of the design. (b-1) A prototype. (b-2,3) The prototype held by a parallel gripper. Note: Only the front end is visible. The back end is the same as the front end except that the rotating direction is reversed.}
    \label{teaser}
    \end{center}
\end{figure}

Under this background, we proposed that instead of directly using the end-effectors, robots may use the mechanical tools to perform various tasks. Over the years, humans were developing tools for themselves. Instead of humans, we proposed to develop tools for intelligent robots. We design mechanical gadget tools considering the constraints from a general robotic gripper and specific tasks. We demonstrate that intelligent robots could select and use the designed tool through vision and planning, and successfully work in narrow workspaces where they cannot reach using their hands. These tools are purely mechanical and free of power supply. They are small and convenient. They can be freely placed in a robot workspace and selected and used by robots. 

In our previous study \cite{zhengtao2019}, we designed a nipper tool for a robot hand to squeeze various sized objects. The nipper enforces a robot gripper by extending the parallel motion. Following a similar conception, we in this paper present a new type of mechanical tool which converts the parallel motion of a robotic gripper to continuous rotation, as is shown in Fig.\ref{teaser}. The tool's essential structure is based on a Scissor-Like Elements (SLEs) and a double-ratchet mechanism. The SLEs form the mainframe of the tool. It helps to keep a fixed rotation center during transmission. In order to make the structure compact, the holding pads are optimized to have a curved profile to reduce the length of the SLE arms. The double-ratchet unit is installed concentrically to the pivoting center of the SLEs. It is made of two ratchets with reversed locking directions. The two ratchets are fixed to the arms of the front and back SLEs respectively. They are connected to each other by a central shaft. Torsion springs are installed at the SLE joints to both provide enough resistant pressure for being held by robotic grippers and stretching the tool. The tool is entirely mechanical, allowing robots to use the tool without any peripherals and power supply. The paper presents the details of the tool design, optimizes its dimensions and effective stroke lengths, and studies the contacts and forces to achieve stable grasping and screwing. 

Besides the design, the paper develops manipulation policies for the tool. The policies include visual recognition, picking-up and manipulation, and exchanging tooltips. The developed tool produces clockwise rotation at the front end and anti-clockwise rotation at the back end. Various tooltips can be installed at both two ends. Robots may employ the developed manipulation policies to exchange the tooltips as well as switching the rotating directions following the needs of specific fastening or loosening tasks. Robots can also reorient the tool using pick-and-place or handover, and move the tool to work poses using the policies. 

The remaining sections of this paper present the design and optimization of the mechanical screwing tool, its manipulation policies, as well as experiments and analysis. Three prototypes of the tool are fabricated and compared to show the design's advantages and its optimization. Experiments are carried out to verify the design. Real-world robotic applications are performed to demonstrate the robustness and usefulness. The tool is small, cordless, convenient, and has good robustness and adaptability.

\section{RELATED WORK}

We concentrate the literature review on screw fastening tasks, robotic end-effector design, and robotic applications that use tools.

\subsection{Screw Fastening Tasks}
The screw fastening is a labor-intensive task faced widely in industrial assembly. A screw fastening task consists of alignment, joining, insertion, and fastening \cite{seneviratne1992theoretical}, which is challenging for robotic manipulation\cite{lee2005studying}. Previous solutions usually use (1) special-purpose robots or special-purpose robotic end-effectors, or (2) general robots plus screwing tools to perform screw fastening tasks.

Lots of specially designed screwing robots are available in the literature. For example, a mobile robot designed for drilling and fastening was presented in \cite{adams2014next}. A 4-DoF screw fastening robot with a visual servo system was presented in \cite{pitipong2010automated}. These robots can effectively complete the fastening job in the structural environment but short in flexibility. Besides the special-purpose robots, many studies use special-purpose end-effectors mounted on the end of a robot arm to perform the screw fastening tasks \cite{cherubini2015unified}\cite{matsuno2013fault}\cite{dhayagude1996fuzzy}. The advantage of a special-purpose end-effector is since it can move more freely and stably following the hosting robot arm. The drawback is an end-effector highly constrains the function of a robot arm into a single-purpose manipulator. Some interesting multi-purpose end-effectors are introduced in \cite{yokokohji2019assembly} to solve the problem, where a robot hand with a built-in screwdriver can both grip and screw, but the mechanism and the various peripheral connections are complex and bulky.

Beyond the special-purpose design, robots can also complete the screw fastening task by using driver tools like humans. For example, an electric driver with a tool adapter for humanoid robots was presented in \cite{pfeiffer2017nut}. The electric driver is a general one, and thus, the system is more flexible, although the tool adapter still needs to be designed carefully. Drigalski et al. \cite{von2020team} and Nie et al. \cite{nie2020team} respectively proposed electric drivers and wrenches which have cuboid bodies for being grasped by general parallel grippers. In \cite{murooka2019self}, a robot is demonstrated to use a hex wrench to fasten the screws of its own body. Developing robots to use tools is an important research topic in robotics. Most previous studies concentrated on the recognition and manipulation of the tools. The relative work to these topics will be reviewed later in the third part of this section.

Unlike using an electric tool or manipulating a driver to perform a screwing motion, we design a mechanical tool which can convert the parallel gripping motion into rotation, and use the tool to fasten screws. Our work involves both the design and manipulation policies for tool use. Thus in the next two subsections, related work on tool design and tool use are presented. Especially since the tool design requires similar knowledge and principles as robotic end-effector design without power supplies, we review the literature in the boarder field of robotic end-effector design.

\subsection{Robotic End-effector Design}
An end-effector is a device mounted on the end of a robot. It is determined according to the usage of a robot, such as suction cups for picking, grippers for grasping, guns and torches for welding, sprayers for painting, sanders for polishing and buffing, etc \cite{miller2013industrial}. Designing a high-performance end-effector is an old and continuing topic in robotics. Previous studies cover a wide range of topics like the mechanical design, the actuation system, and the control of robotic end-effectors \cite{nof1999handbook}\cite{monkman2007robot}\cite{wolf2018grippers}. Since our goal is to design a mechanical tool for screwing, we focus on the literature about the mechanical structure. Our expectation is when the tool is held by a gripper, it can continuously output rotating torque in accordance with the continuous close-and-open of the gripper. Scissor-Like Element (SLE) and ratchet mechanism are employed as basic elements for the tool. Elastic elements are used to provide resisting force for holding the tool and producing torque output when the gripper releases the tool. We consequently focus our review on the SLE mechanism, the ratchet mechanism, and the elastic elements below.

SLE is a widely seen mechanical unit in scissors and scissor-like tools like pliers. A basic SLE element has two scissor arms that can freely rotate around a pivoting point \cite{zuk1970kinetic}. This basic element has many variations. For example, Monkman et al. \cite{monkman2007robot} and Khasawneh et al. \cite{khasawneh2014enhanced} respectively extended basic SLEs to a pantograph for transmitting the grasp stroke of grippers. Maden et al. \cite{maden2011review} reviewed Chained SLEs (C-SLE) used in planar or spatial structures. The  C-SLE is a popular mechanism for robotic end-effector design. Yang et al. \cite{yang2017novel} presented a 2-DoF planar translational mechanism based on SLE-parallel -- a mechanism consisting of two identical SLE limbs which are connected at the two corresponding nodes by link. Corinaldi et al. \cite{corinaldi2016synthesis} proposed 3-DoF deployable gripper mechanism using SLEs and Sarrus linkages, which has a spatial structure to transmit motion symmetrically. Kocabas et al. \cite{kocabas2009gripper} developed a 1-DoF spherical gripper mechanism consisting of spiral SLEs and linkages for power grasping of various shapes. Ndawula et al. \cite{ndawula2018conceptual} proposed a double planar scissor mechanism mounted with multi-gripper for handling a row of seedlings. Not only the robotic end-effector designing, but SLE is also widely used in the structure of robot bodies \cite{bamdad2015design}\cite{luo2016scissor}, exoskeleton \cite{castro2019compact}, as well as more general mechatronic devices to perform tasks like mobile pavilions, foldable stairs, collapsible doors, etc \cite{zhao2011structure}\cite{akgun2011novel}.

A ratchet is a mechanism that allows continuous linear or rotary motion in one direction but locks the motion in the opposite direction. The feature makes a ratchet a widely used transmission mechanism \cite{rao2014power}\cite{dao2008micro}. In robotics, a ratchet is usually used as a locking device \cite{plooij2015lock}. Li et al. \cite{li2009spherical} developed a hopping robot, in which a locked ratchet mechanism is released to trigger an energy storage mechanism. Geeroms et al. \cite{geeroms2013ankle} developed an active knee-ankle prosthesis, in which a ratchet unit is used to lock the weight acceptance mechanism. As for the end-effector design, Abe et al. \cite{abe2012development} designed a reconfigurable end-effector for endoscopic surgery by using a bending mechanism where a ratchet unit is employed to lock and release transmission following the bending conditions. Gerez et al. \cite{gerez2018compact} and Sabetian et al. \cite{sabetian2011compound} focused on the development of underactuated grippers using ratchets. Besides, an electrostatic microgripper was presented in \cite{hao2015microgripper} by taking advantage of the locking feature of a ratchet.

Elastic elements are usually used for passive actuation in robotic mechanism design. Laliberte et al. \cite{laliberte2002underactuation} designed an underactuated finger by using elastic elements as switches for the underactuated joints. Xu et al. \cite{xu2016continuum} applied continuum differential mechanisms to a gripper design, in which elastic elements are used as bendable backbones. Chen et al. \cite{chen2018mechanical} analyzed the adaptability of the underactuated grippers constrained by elastic elements.

Instead of an end-effector, we design a tool for 2-finger parallel robotic grippers. The actuation force of the tool is only from the robotic grippers. Two modified C-SLE (mC-SLE) are used to convert the parallel robotic gripping motion into oscillating rotation. A double-ratchet mechanism is then connected to the two mC-SLEs to further convert the oscillating rotation into a unidirectional rotation. Like our previous nipper tool \cite{zhengtao2019}, elastic elements are used to provide resisting force for robotic grasping. They are also used to stretch the tool and maintain rotation when a robotic gripper releases the tool. 

\subsection{Robotic Manipulation of Tools}
Using tools is an extensively studied robotic manipulation problem. With known models, developing robotic applications to use tools can be formulated and solved as a logical reasoning problem \cite{toussaint2018differentiable}\cite{brown2012relational}. The motion for using the tool can be planned by combined task and motion planning \cite{chen2019}. The task routine for using a tool is complicated, which, however, can be resolved into several subproblems \cite{gupta1998micro}\cite{tikhanoff2013exploring} like tool selection, tool recognition, constrained grasping and tool reorientation, etc. \cite{saito2018detecting}\cite{fang2020learning}\cite{saito2018tool}\cite{li1988task}. Practical systems can thus be implemented in a divide-and-conquer way. 
Learning from demonstrations is also a popular approach to transfer the routine of using tools to robots. In \cite{Rajeswaran-RSS-18}, robots learned complicated manipulation like using a hammer through simulated demonstrations. In \cite{zhu2015understanding}, human demonstrations were captured by pose-tracking and were then employed to learn how to identify and use tools. Raessa et al. \cite{raessa2019teaching} proposed a human demonstration-based method for teaching robots to use tools with special consideration of regrasp planning. Additionally, instead of demonstration, Xie et al. \cite{xie2019improvisation} presented the method of using video prediction for reasoning the potential robot action to use the available objects as tools in an improvisational way. Besides the motion and task routine, many studies also focus on the various constraints and force problems. For example, Rachel et al. \cite{holladay2019force} studied the force constraints in the tool manipulation tasks. Toussaint et al. \cite{toussaint2018differentiable} studied the physical interaction between a tool and an object. 

In this work, we design the mechanical screwing tool and develop the manipulation policies for a robot to use the tools for screw fastening tasks. Our novelty in using the tool is two-fold. First, the tool is a mechanical one with links, pivoting joints, and elastic elements. Thus we have to carefully study the contacts and forces for stable grasping and use. Second, our manipulation policies include recognition, picking-up and manipulation, and exchanging tooltips. The manipulation policies, especially exchanging the tooltips, extends the flexibility of the tool and the robotic system.

\section{THE MECHANICAL STRUCTURE}

\subsection{Scissor-Like Elements (SLEs)}
\subsubsection{The basic SLE and its problems}
\begin{figure}[!htbp]
    \begin{center}
    \includegraphics[width=0.97\linewidth]{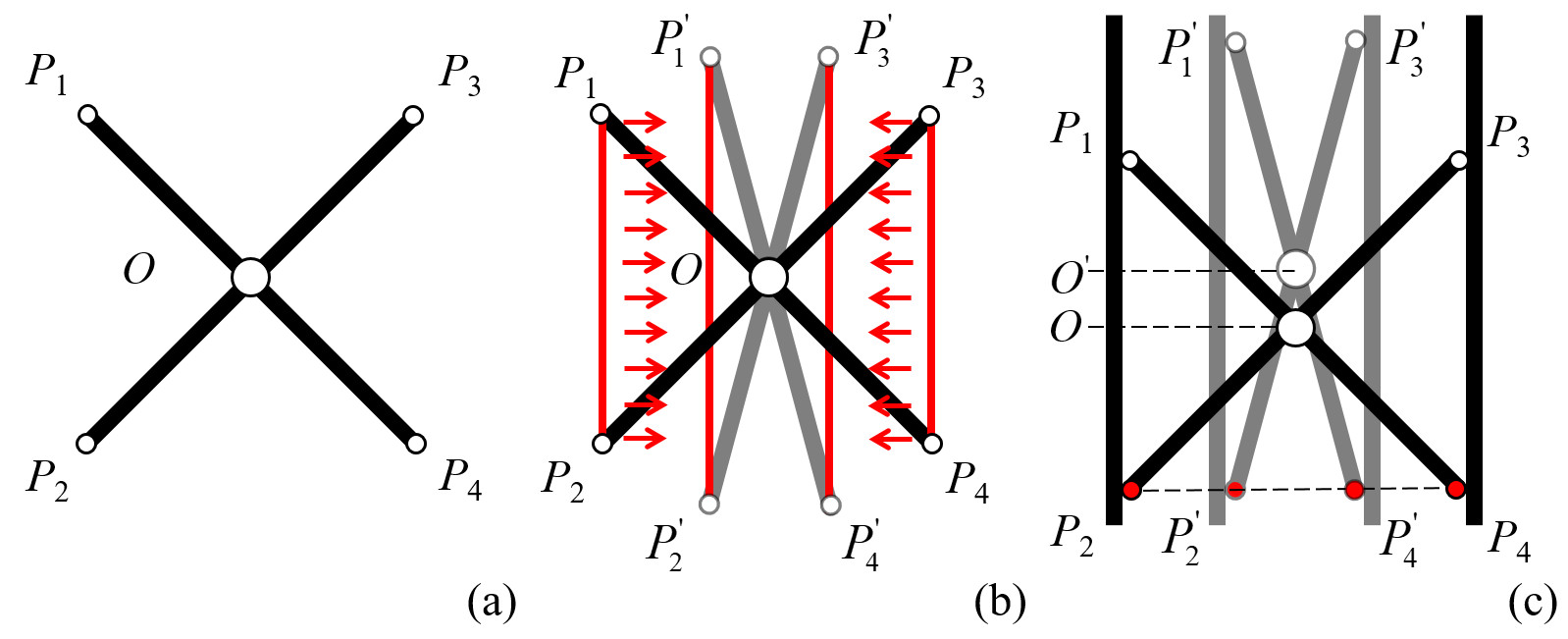}
    \caption{(a) A basic Scissor-Like Element (SLE). (b) The rotation of a basic SLE is around its pivoting joint. The rotation is converted into the translation motion of $segment$s $P_{1}P_{2}$ and $P_{3}P_{4}$, which could potentially afford the motion of a parallel gripper. However, the length of $segment$ $P_{1}P_{2}$ (also $segment$ $P_{3}P_{4}$) changes during the translation, making them not suitable to be held by fingers. (c) An intuitive solution to implement two graspable pads. $P_{2}$ and $P_{4}$ are rotating joints that connect the SLE arms and the two pads. $P_{1}$ and $P_{3}$ are free ends. The two pads keep steady as they close. A problem of this intuitive solution is the pivoting joint $O$ moves with the free ends.}
    \label{basicsle}
    \end{center}
\end{figure}

Fig.\ref{basicsle}(a, b) show a basic SLE. The rotation of the basic SLE induces the translational motion of $segment$ $P_{1}P_{2}$ and $segment$ $P_{3}P_{4}$ (see the red segments and the arrows in Fig.\ref{basicsle}(a-b)). This property makes the basic SLE a good candidate mechanism for our tool, for the linear motion of $segment$ $P_{1}P_{2}$ and $segment$ $P_{3}P_{4}$ affords the motion of a parallel gripper.

A shortcoming of the basic SLE is the length of $segment$ $P_{1}P_{2}$ (also $segment$ $P_{3}P_{4}$) changes during the translation. They turn into longer $segment$s $P^{'}_{1}P^{'}_{2}$ and $P^{'}_{3}P^{'}_{4}$, making them not suitable to be held by fingers. Thus, to ensure a stable hold, some points on the SLE arms must be fixed. Fig.\ref{basicsle}(c) shows an intuitive solution used in mini scissor lift. In this case, $P_{2}$ and $P_{4}$ are attached to two pads using rotational joints. $P_{1}$ and $P_{3}$ are free ends. As the two pads move close, $P_{2}$ and $P_{4}$ move up to $P^{'}_{2}$ and $P^{'}_{4}$. The pads move in a stable linear motion, which is suitable to be held.

The intuitive solution enables stable and linear pad motion but still does not meet the requirements of a screwing tool. As the pads close, the pivoting joint $O$ also moves up to $O^{'}$, as shown in Fig.\ref{basicsle}(c). A changing pivoting joint is not a good candidate for attaching the tooltip of a screwing tool. Thus further modification is needed to make the basic SLE valuable. 

\subsubsection{Chained SLE (C-SLE) and the proposed modification}
\begin{figure}[!htbp]
    \begin{center}
    \includegraphics[width=0.97\linewidth]{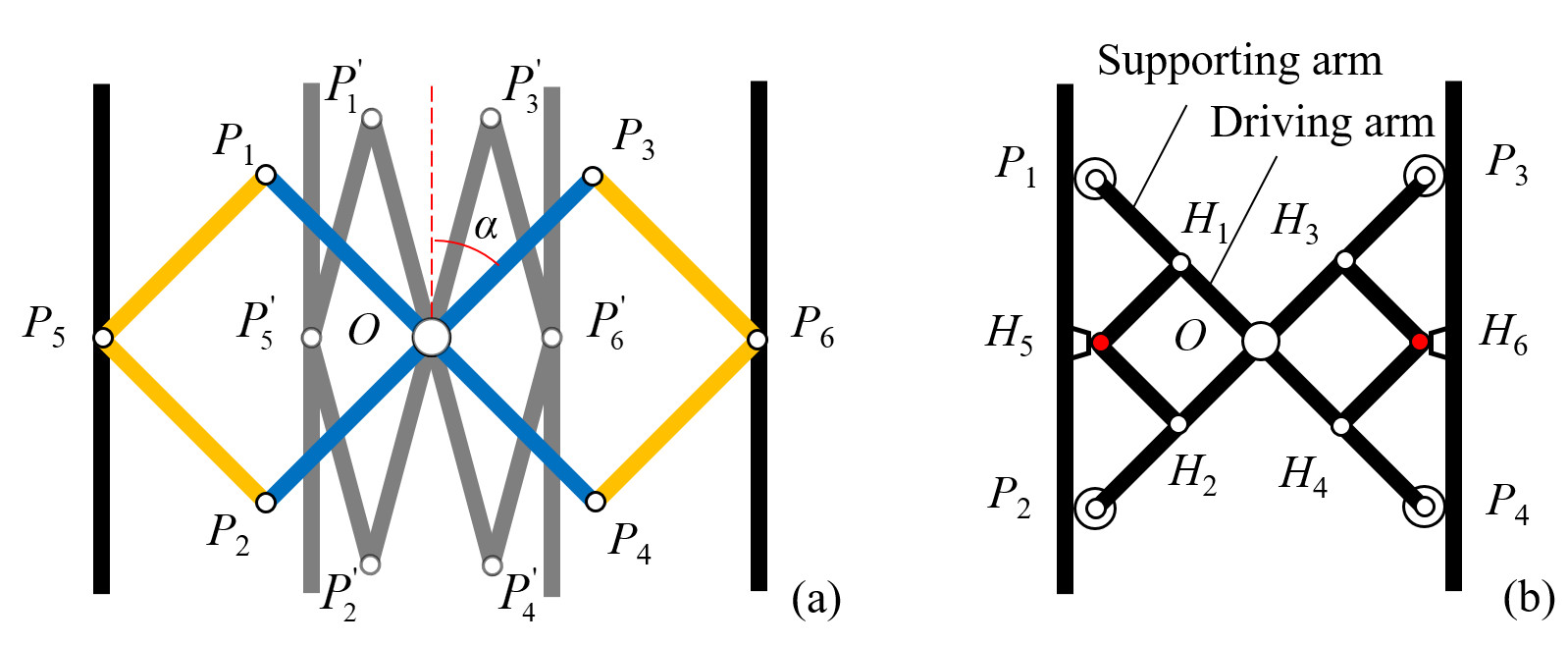}
    \caption{(a) The structure and motion of a kinematic SLE chain made of one full SLE (blue) and two half SLEs (yellow). While $P_{5}$ and $P_{6}$ get close, $\alpha$ will increase so that the arms will rotate around $O$. Compared to Fig.\ref{basicsle}(c), the position of $O$ remains unchanged in the structure. The disadvantage of this design is that it cannot withstand external torque as there are only rotational constraints at $P_{5}$ and $P_{6}$. (b) The proposed C-SLE (Chained SLE) design. In this design, the arms of $OH_i (i=1,2,3,4)$ are further extended to $P_{i}$ to resist torque.}
    \label{chainedsle}
    \end{center}
\end{figure}

We propose to use a modifiend C-SLE to solve the problems mentioned above. Following the C-SLE concept, one full and two half SLEs can be chained together to keep the position of the pivoting joint fixed while the two pads close. Fig.\ref{chainedsle} illustrates the idea. In Fig.\ref{chainedsle}(a), the blue links form a full SLE. The yellow links form two half SLEs. $P_{5}$ and $P_{6}$ are the rotational joints connected to the holding pads. While the gripper squeezes the mechanism, the distance between $P_{5}$ and $P_{6}$ will decrease so that $\alpha$ will be increased to make the arms rotate around the pivoting joint $O$. Also, $O$ is the center of $segment$ $P_{5}P_{6}$, and $P_{5}$-$O$-$P_{6}$ keep collinear all the time. The position of $O$ will remain unchanged while the robot gripper presses against $P_{5}$ and $P_{6}$.

The disadvantage of the configuration shown in Fig.\ref{chainedsle}(a) is that it cannot withstand an external torque as there are only rotational constraints at $P_{5}$ and $P_{6}$. The design collapses if the pressing forces are not exactly exerted at $P_{5}$ and $P_{6}$. To avoid this problem, we extend the arms of the full basic SLE for support. As shown in Fig.\ref{chainedsle}(b), the cross $H_1H_4$-$H_2H_3$ is the full SLE. The $H_1$-$H_5$-$H_2$ and $H_3$-$H_6$-$H_4$ on the two sides are the two half SLEs. By extending $OH_1$ to $P_1$, $OH_2$ to $P_2$, $OH_3$ to $P_3$, and $OH_4$ to $P_4$, and use $P_1$-$P_4$ as the supporting joints, the mechanism can accept pressure at any positions on the two holding pads. The parallel motion of a robot gripper can be converted into a rotational motion around $O$. $P_{1}$-$H_{5}$-$P_{2}$ and $P_{3}$-$H_{6}$-$P_{4}$ maintain collinear during the rotation. Note that all the $P_{i} (i=1,2,3,4)$ play the role of a free end. Supporting wheels are therefore installed on them to enable the free motion. The arms $H_1P_1$, $H_2P_2$, $H_3P_3$, $H_4P_4$ are called the supporting arms. They are symmetric and have the same length. The arms $OH_1$, $OH_2$, $OH_3$, $OH_4$ are called the driving arms. They are also symmetric and have the same length.

This modified C-SLE design is called mC-SLE. We install two mC-SLEs in parallel to hold the double-ratchet mechanism and realize continuous rotating motion. The details are presented in the next subsection.

\subsection{Double-Ratchet Mechanism}
\subsubsection{The basic ratchet and its problems}
\begin{figure}[!htbp]
    \begin{center}
    \includegraphics[width=0.48\textwidth]{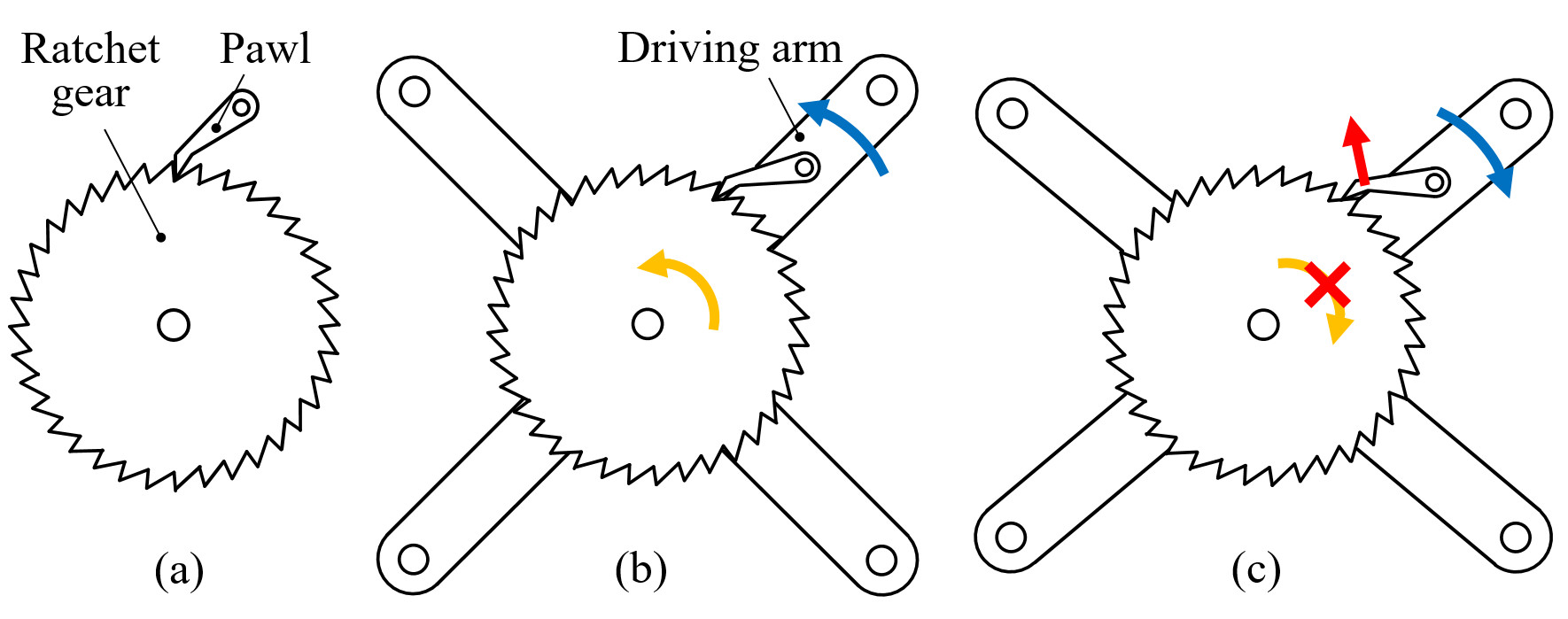}
    \caption{(a) The sketch of a basic ratchet. (b, c) The pawl of the ratchet is fixed on a driving mC-SLE arm. (b) The locking condition. In this condition, a mC-SLE arm drives the ratchet gear to rotate anti-clockwise. (c) The released condition. In this condition, the pawl is pushed up by the ratchet gear, and the gear does not rotate with the motion of the pawl.  Note that the illustration in (c) is not always true. When the friction between the pawl tip and the gear teeth is larger than the gear's rotational resistance, the gear may rotate back together with the pawl, leading to a failure in the single-direction regulation.}
    \label{basicratchet}
    \end{center}
\end{figure}

At the pivoting joint of the mC-SLE, as shown by the blue arrows in Fig.\ref{basicratchet}(b, c), the rotating motion oscillates with the open and close of the holding pads. Such motion does not meet the requirements of a screwing tool, where only a single-direction motion is needed. Thus, we use a ratchet mechanism to regulate the oscillating rotation into a single-direction rotation. Fig.\ref{basicratchet}(a) illustrates a basic ratchet. It comprises a ratchet gear or a rack with teeth and a pawl engaged with the gear teeth for locking. We attach the gear and the pawl of the basic ratchet to the pivoting center and one arm of the mC-SLE respectively. The arm will drive the pawl to push the ratchet gear to rotate in a locked condition, as shown by the yellow arrow in Fig.\ref{basicratchet}(b). In a released condition, as shown in Fig.\ref{basicratchet}(c), the pawl is pushed up by the ratchet gear and the rotation stops.

A problem with the attachment is the ratchet gear does not necessarily stay stationary in a released condition. If the friction between the pawl tip and the gear teeth is larger than the rotational resistance of the gear, the gear may rotate back together with the pawl, leading to a failure in the single-direction regulation. Although a heavy rotation load may provide enough external torque to overcome the friction, it is, however, not a reasonable assumption as not all objects provide heavy load.

Also, even if the external torque in the released condition is large enough, the resulted motion is still an intermittent rotation instead of a continuous one. Thus, some modification to the basic ratchet must be included.

\subsubsection{The proposed double-ratchet mechanism}
\begin{figure}[!htbp]
    \begin{center}
    \includegraphics[width=0.48\textwidth]{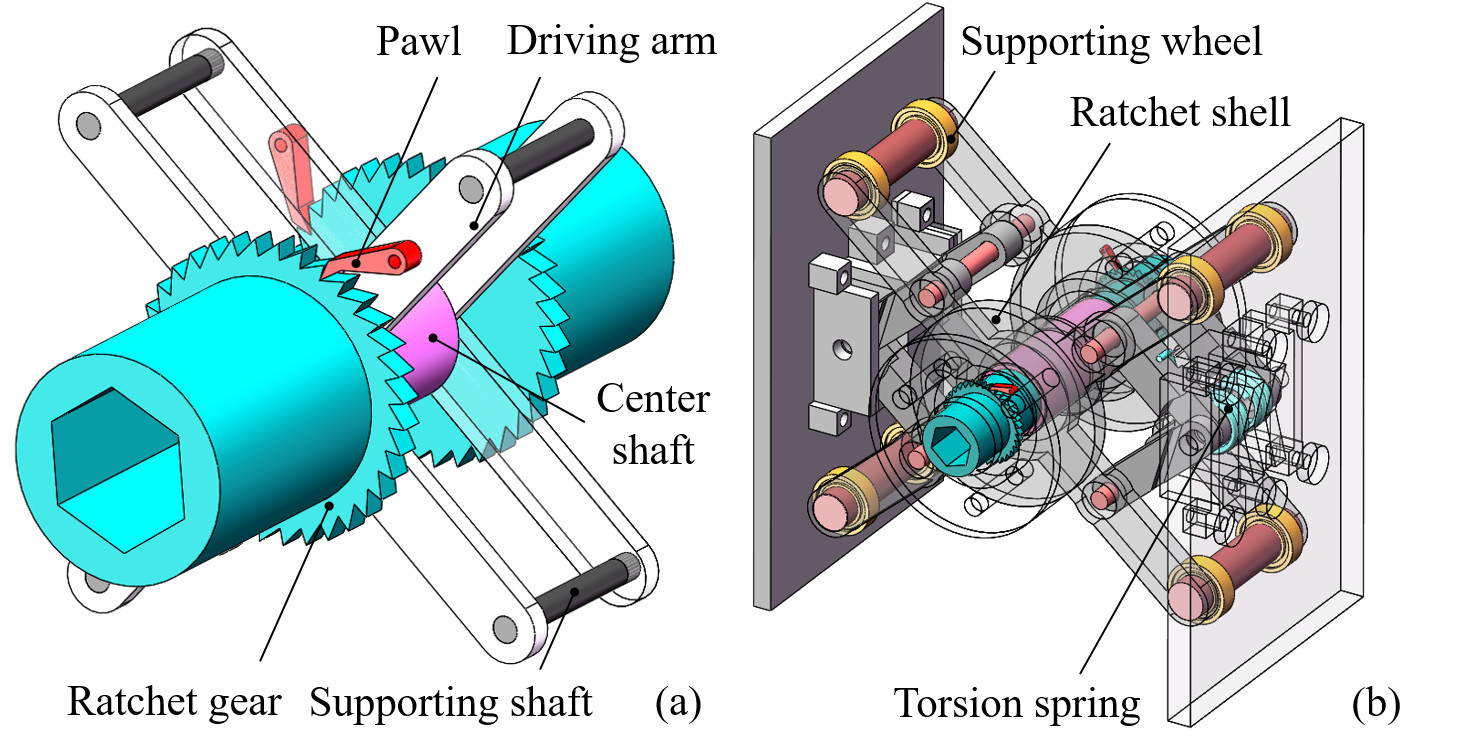}
    \caption{(a) A double-ratchet mechanism driven by inversely rotating arms. (b) The proposed mechanical screwing tool design using two C-SLEs and a double-ratchet mechanism. The design is capable of producing continuous rotation.}
    \label{doubleratchet}
    \end{center}
\end{figure}

We propose using a double-ratchet mechanism to solve the aforementioned problems. The design is shown in Fig.\ref{doubleratchet}(a). The pawls of the two basic ratchets are attached to two inversely rotating arms. When the front ratchet is in a locked condition, the back ratchet in a released condition. The front gear and also the central shaft can be driven by the front pawl to rotate in clockwise. The back gear also moves clockwise together with the central shaft. The friction between the pawl tip and the gear teeth of the bear ratchet mechanism is overcome by the driving force of the front ratchet. When the front ratchet is in a released condition, the back ratchet is in a locked condition. It will rotate clockwise and drive the central shaft and the gear of the front ratchet to rotate in the same way. Continuous clockwise rotation can thus be realized.

Fig.\ref{doubleratchet}(b) illustrates a mechanical screwing tool designed considering both the aforementioned mC-SLE and double-ratchet mechanism. The design can output rotation continuously. It involves two parallel and colinear mC-SLEs. On the front mC-SLE, a ratchet is attached to perform clockwise rotation when the pads are closed. Another ratchet is attached to an arm on the back mC-SLE to continue the rotation when the pads are open. The two ratchets are attached to inversely rotating arms to make sure they perform a rotation in a different direction when locked. 

\section{ANALYSIS AND OPTIMIZATION}

In this section, we perform quasi-static force analysis on the design to study and optimize the output of the tool, as well as to make the tool stable and compact.  

\subsection{Grasping the Tool}
\subsubsection{The condition to stably hold the tool}
We formulate the contact between the robot fingers and the holding pads as soft finger contacts \cite{harada2014stability}. The friction and force at the contact meet the following equation: 
\begin{equation}
f_{grpr}^{2}+\frac{T_{grpr}^{2}}{e^{2}}\leqslant \mu^{2}P_{grpr}^{2}\label{eq-fa-1},
\end{equation}
where $P_{grpr}$ is the pressure force exerted by the robot finger, $T_{grpr}$ and $f_{grpr}$ are the torque and tangential force, $\mu$ is the friction coefficient, $e$ is an eccentricity parameter which can be computed as:
\begin{equation}
    e^{2}=\frac{\max T_{grpr}^{2}}{\max f_{grpr}^{2}}\label{eq-fa-2}.
\end{equation}

\begin{figure}[!htbp]
    \begin{center}
    \includegraphics[width=0.48\textwidth]{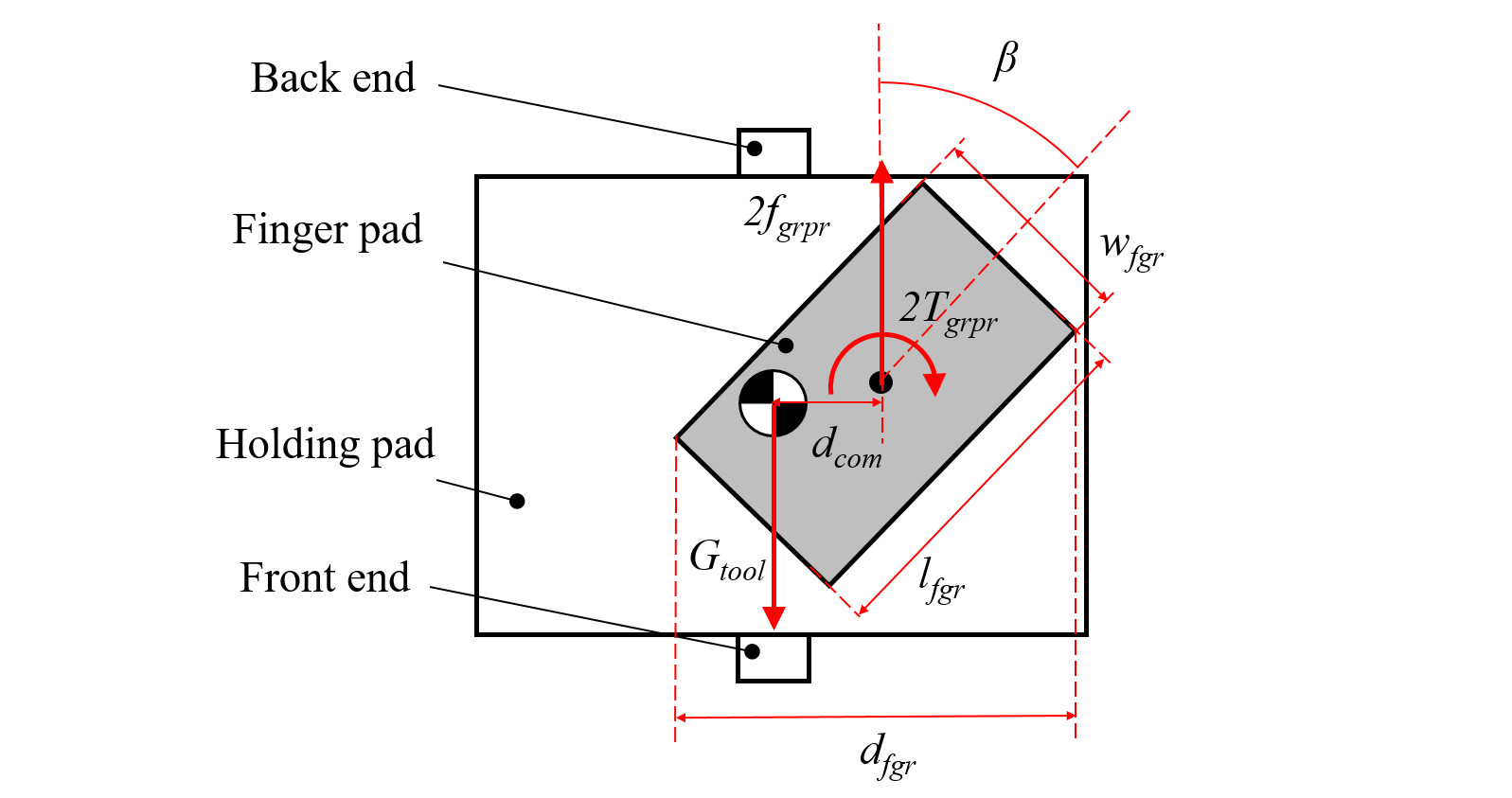}
    \caption{The free body diagram of the tool from a side view.}
    \label{fbd-sideview}
    \end{center}
\end{figure}

\begin{figure}[!htbp]
\begin{center}
\includegraphics[width=0.48\textwidth]{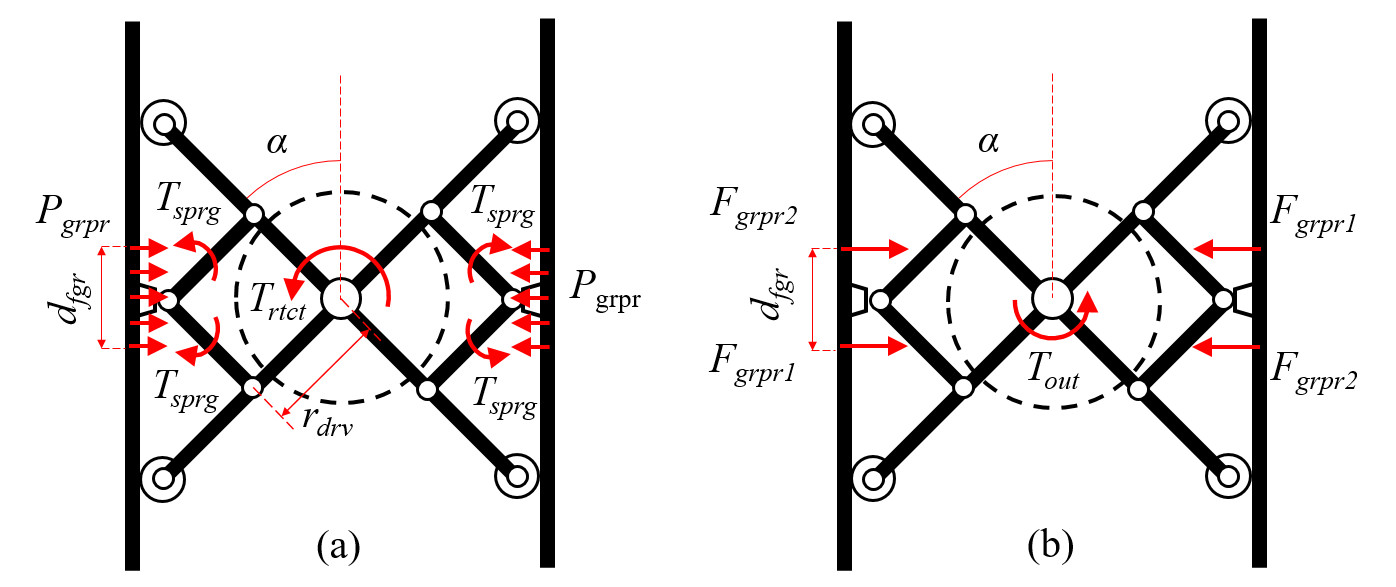}
\caption{(a) The free body diagram of the tool when it is held by a robot gripper. (b) The freebody diagram of the tool when it is bearing a maximum rotational load.}
\label{fbd-vertical}
\end{center}
\end{figure}

When grasping the tool, the robotic finger pad may contact with the tool holding pad at an arbitrary position. Fig.\ref{fbd-sideview} shows an example. Here, $d_{com}$ indicates the distance between the Center of Mass (CoM) of the tool and the center of the contact. The forces at the contact can be analyzed following two conditions, as shown in Fig.\ref{fbd-vertical}. The first one is when the robot grasps the tool sufficiently. In this case, the force at the contact is a distribution shown in Fig.\ref{fbd-vertical}(a). The contact position must meet the following equation to make sure the tool can be stably held:
\begin{equation}
    d_{com}\leqslant\sqrt{\frac{4\mu^{2} P^{2}_{grpr}e_{grpr}^{2}}{G_{tool}}-e^{2}}.
    \label{condition}
\end{equation}
Here, $P_{grpr}$ can be computed by
\begin{equation}
    P_{grpr}=\frac{4T_{sprg}+T_{rtct}}{4r_{drv}\cos{\alpha}},
\end{equation}
and
\begin{equation}
    T_{sprg}=\xi(\gamma+\alpha-\alpha_{init}),
\label{springco}
\end{equation}
where $\xi$ is the elasticity coefficient of the torsion springs ($N\cdot mm/^{\circ}$). $\gamma$ indicates the pre-set rotational deformation of the torsion springs used in assembling the tool. $\alpha$ is the angle between the mC-SLE's driving arm and the holding pad. $\alpha_{init}$ is the $\alpha$ angle when the tool is at a free (initial) state. $r_{drv}$ is the length of the mC-SLE's driving arm. $T_{rtct}$ is the resisting torque caused by a reversing pawl.

The second one follows a critical condition where the contact forces concentrate at one side of the finger pads. It happens when the tool is bearing a maximum rotational load. The forces must also meet \eqref{condition} to firmly hold the tool. Compared to the first one, the second case allows us to analyze the extremes, namely the maximum output torques at the tooltip. Thus we are more interested in it and analyze it in detail.

\subsubsection{Maximum output torque at the tooltip}
The critical condition in the second case has two extremes. The first extreme appears when the tool holding pads are squeezed. At this extreme, the output ends of the tool are rotated by the grabbing force of the robot gripper. The second extreme appears when the tool holding pads get stretched freely. At this extreme, the output ends are rotated by the repulsive force of the compressed torsion springs. The output torque at the tooltip reaches the squeezing maxima and the stretching maxima at the two extremes. Assume the grabbing force is $F_{grpr}$. At the first extreme, $F_{grpr}$ concentrates at $F_{grpr1}$, while $F_{grpr2}=0$. The maximum output torque can be computed by:
\begin{equation}
    T_{sqz}=(F_{grpr}-\frac{T_{sprg}}{r_{drv}\cos{\alpha}})d_{fgr},
\label{output_sqz}
\end{equation}
where 
\begin{equation}
    d_{fgr}=w_{fgr}\sin{\beta} + l_{fgr}\sin{\beta}.
\label{output-dfinger}
\end{equation}
The symbols in equation \eqref{output-dfinger} are illustrated in Fig.\ref{fbd-sideview}. $w_{fgr}$ and $l_{fgr}$ are the width and height of the holding finger pad respectively. $\beta$ is the holding angle.

At the second extreme, $F_{grpr}$ concentrates at $F_{grpr2}$, while $F_{grpr1}=0$. The maximum output torque can be computed by:
\begin{equation}
    T_{stch}=\frac{T_{sprg}}{r_{drv}\cos{\alpha}}d_{fgr}.
\label{output_stch}
\end{equation}

\begin{figure}[!htbp]
\begin{center}
\includegraphics[width=0.48\textwidth]{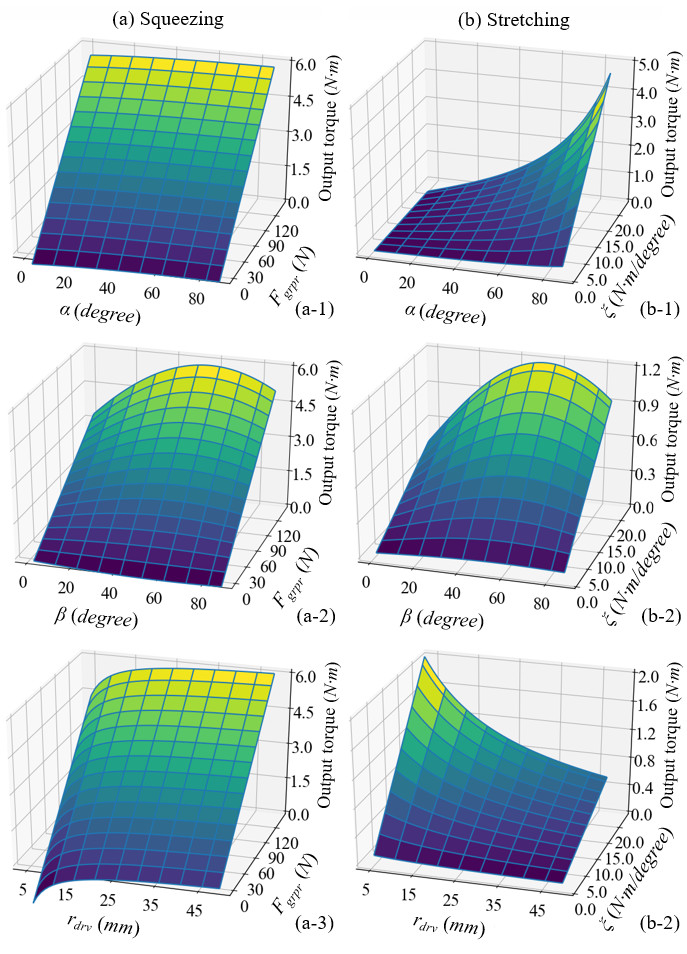}
\caption{The left and right columns respectively show the changes of the output torque with respect varying $F_{grpr}$, $\alpha$, $\beta$, $r_{drv}$ in the squeezing phase, and varying $\xi$, $\alpha$, $\beta$, $r_{drv}$ in the stretching phase.}
\label{squeezingplot}
\end{center}
\end{figure}

To have an intuitive view of the relation between $F_{grpr}$ and $T_{sqz}$, we set the parameters $\gamma$, $\xi$ to constants ($\gamma\leftarrow$0$^{\circ}$, $\xi\leftarrow$6.00$N\cdot mm/^{\circ}$) and examine the changes of $T_{sqz}$ with repsect to varying $F_{grpr}$, $\alpha$, $\beta$, and $r_{drv}$. The changes show insights on how the grasping force, jawwidth, and grasping pose influence the squeezing output. Fig.\ref{squeezingplot}(a) plots the results. Here, $\alpha$, $\beta$ and $r_{drv}$ are decoupled to make the function map visualizable. In Fig.\ref{squeezingplot}(a-1), $\beta$ and $r_{drv}$ are set to 57$^\circ$ and 20$mm$ to visualize the relation between ($F_{grpr}$, $\alpha$) and $T_{sqz}$. Fig.\ref{squeezingplot}(a-2) shows the changes of $T_{sqz}$ with repsect to different $F_{grpr}$ and $\beta$. $\alpha$ and $r_{drv}$ are set to fixed values, i.e., 45$^{\circ}$ and 20$mm$. The influence of $F_{grpr}$ and $\beta$ on $T_{sqz}$ is shown in Fig.\ref{squeezingplot}(a-3). Here, $\alpha$ and $\beta$ are fxied to 45$^{\circ}$ and 57$^{\circ}$.

Likewise, for the stretching output, Fig.\ref{squeezingplot}(b) plots changes of $T_{stch}$ with respect varying $\xi$, $\alpha$, $\beta$, and $r_{drv}$. $\gamma$ is set to $0^{\circ}$ in the analysis. The other parameters are fixed in the same way as the analysis in the squeezing phase.

\subsubsection{Velocities at the tooltip}
\begin{figure}[!htbp]
\begin{center}
\includegraphics[width=0.48\textwidth]{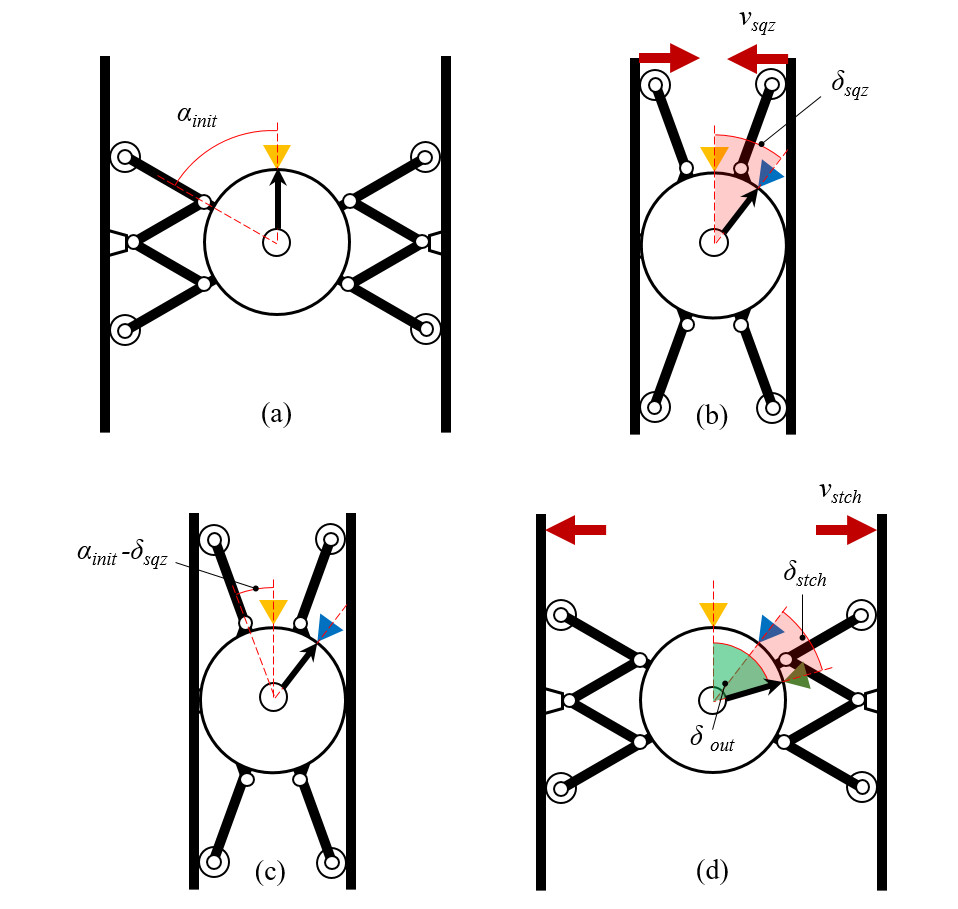}
\caption{(a) The initial angle in a squeezing phase. (b) The output angle at the end of a squeezing phase. (c) The initial angle of a stretching phase. (d) The output angle at the end of a stretching phase.}
\label{output-direction}
\end{center}
\end{figure}

The rotation of the tooltips continuously spans the squeezing and stretching phases, as is shown in Fig.\ref{output-direction}. The black arrow in the figure marks the rotation of the output end. The triangle markers indicate the rotating angle from the starting position. Fig.\ref{output-direction}(a, b) is the squeezing phase. In this phase, the black arrow points to the yellow triangle marker initially. As the tool gets squeezed, the output end rotates clockwise to the blue triangle marker. The angle between the blue and yellow triangle markers is the squeezing phase's output rotation angle, which can be expressed as:
\begin{equation}
    \delta_{sqz}=\alpha_{init}-\sin^{-1}{\left(\sin{\alpha_{init}}-\frac{v_{sqz}\cdot t_{sqz}}{4r_{drv}}\right)},
\end{equation}
where $\delta_{sqz}$, $t_{sqz}$ and $v_{sqz}$ indicate the output angle, the operating time, and the velocity of the gripper fingers in the squeezing phase, respectively.

Fig.\ref{output-direction}(c, d) is the stretching phase that follows (a, b). The black arrow points to the blue marker initially. It is at the end state of the squeezing. As the tool gets stretched, the output end continues to rotate clockwise to the green triangle marker. The angle between the green and blue triangle markers is the output rotation angle of the stretching phase, which can be expressed as:
\begin{equation}
    \delta_{stch}=-(\alpha_{init}-\delta_{sqz})+\sin^{-1}{\left(\sin{(\alpha_{init}-\delta_{sqz})}+\frac{v_{stch}\cdot t_{stch}}{4r_{drv}}\right)},
\end{equation}
where $\delta_{stch}$, $t_{stch}$ and $v_{stch}$ indicate the output angle, the operating time, and the velocity of gripper fingers in the stretching phase.

\begin{figure}[htbp]
\begin{center}
\includegraphics[width=0.48\textwidth]{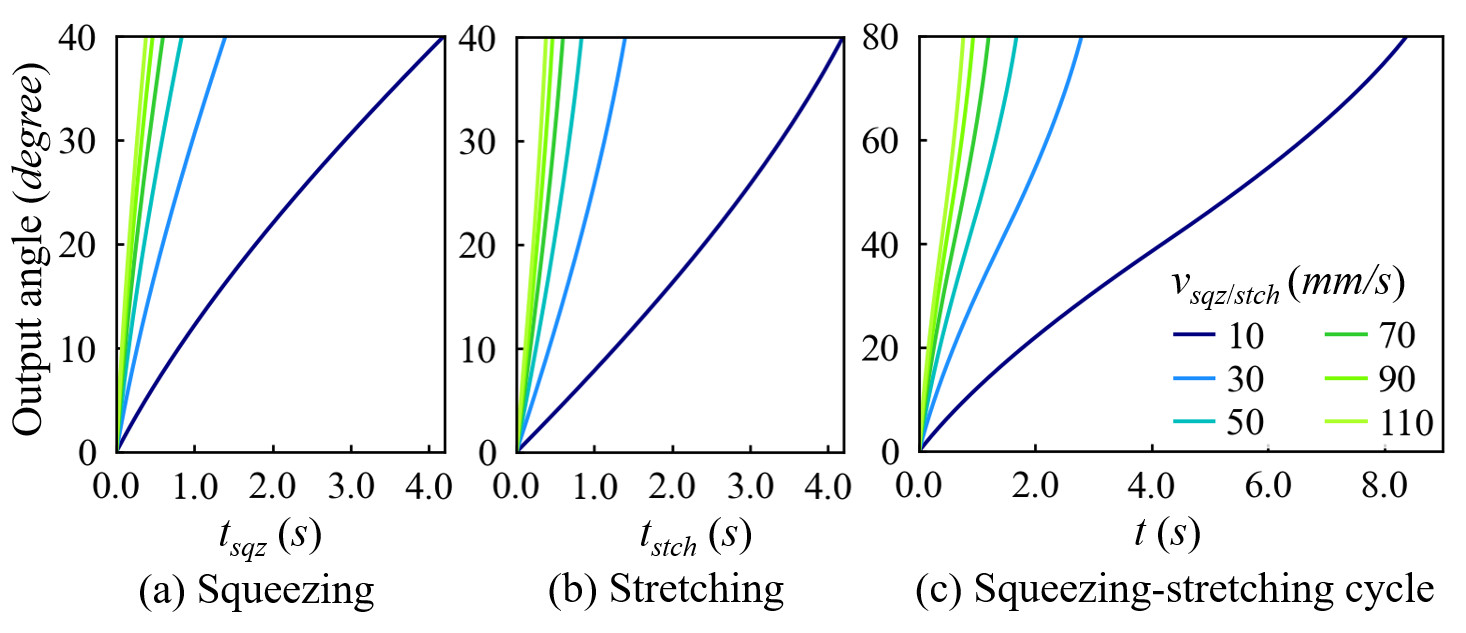}
\caption{The changes of the output angle with respect to $t$ and different $v_{sqz}$$_/$$_{stch}$. The curves with different colors compare different finger speeds. (a) The changes in a squeezing phase. (b) The changes in a stretching phase. (c) The changes in a whole squeezing-stretching cycle. $v_{sqz}$ is set to be equal to $v_{stch}$ in computing the curves in a whole cycle.}
\label{output-angle}
\end{center}
\end{figure}

For a whole squeezing-stretching cycle, the total rotation angle, $\delta_{out}$, equals to the sum of $\delta_{sqz}$ and $\delta_{stretch}$. It is represented by a green sector in Fig.\ref{output-direction}(d). $\delta_{out}$ can be represented as a piecewise function of $t$ as follows
\begin{equation}
\begin{aligned}
    \delta_{out}(t)=\left\{
    \begin{array}{lcr}
\alpha_{init}-\sin^{-1}{\left(\sin{\alpha_{init}}-\frac{v_{sqz}\cdot t}{4r_{drv}}\right)} &     & {t \leq t_{m}},\\
 &     & \\
\alpha_{init}-2\sin^{-1}{\left(\sin{\alpha_{init}}-\frac{v_{sqz}\cdot tm}{4r_{drv}}\right)} &        & \\
+\sin^{-1}{\left(\sin{\alpha_{init}}-\frac{v_{sqz}\cdot (t-t_{m})}{4r_{drv}}-\frac{v_{stch}\cdot (t-t_{m})}{4r_{drv}}\right)} &        & {t > t_{m}}.
    \end{array} \right.
\end{aligned}
\end{equation}
Here, $0\leq t\leq t_m$ is the squeezing time interval, $t>t_m$ is the ensuing stretching time interval. The tool is squeezed to the extreme at $t_{m}$. After that, the stretching phase gets started. Fig.\ref{output-angle} shows the curves of these two equations under different $v_{sqz}$$_/$$_{stch}$ values. The range is from 10 to 110$mm/s$ with a 10$mm/s$ step length. The curve corresponding to each of these values is rendered in a different color. The results show that an increased $v_{sqz}$$_/$$_{stch}$ significantly reduces the operating time.

\subsection{Selecting Proper Torsional Springs}
We choose the torsional springs considering the balance of the output torques in both squeezing and stretching phases. The reversed signs of $T_{sprg}$ in equations \eqref{output_sqz} and \eqref{output_stch} imply a trade-off between $T_{sqz}$ and $T_{stch}$: $T_{sqz}$ monotonically increases with $T_{sprg}$. Contrarily, $T_{stch}$ monotonically decreases with $T_{sprg}$. The spring forces are resistant in the squeezing phase but propulsive in the stretching phase. A larger spring coefficient will lead to large output torque in the stretching phase, but a weaker one in the squeezing phase. 

\begin{figure}[!htbp]
\begin{center}
\includegraphics[width=0.48\textwidth]{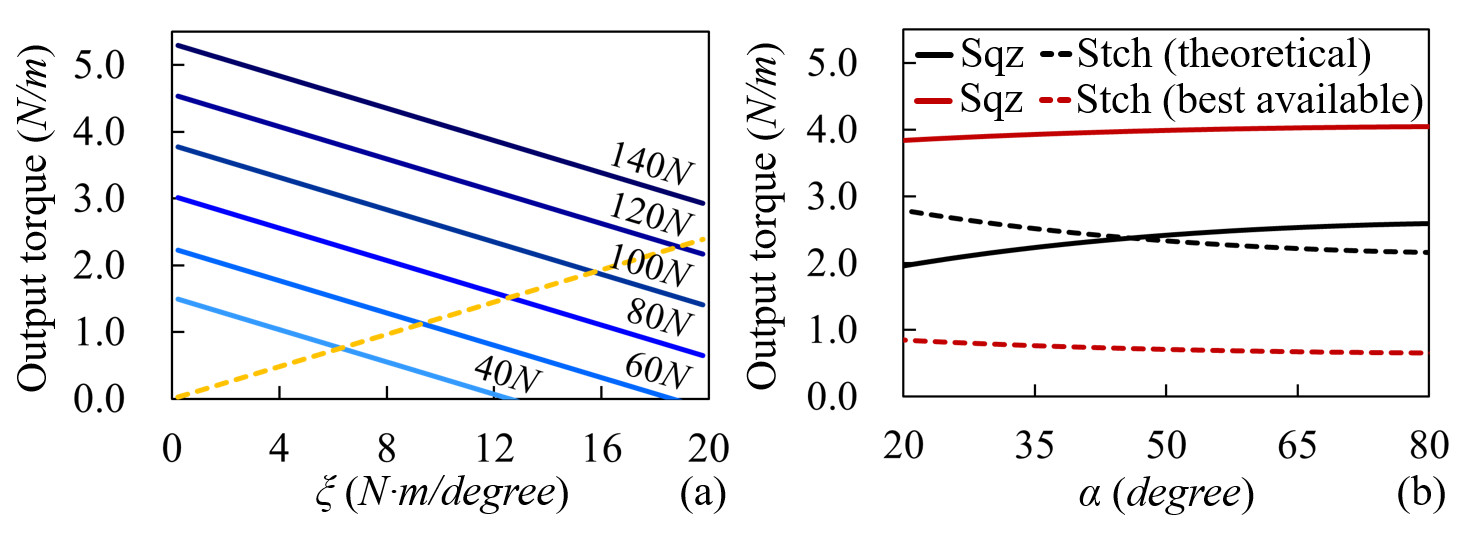}
\caption{(a) Blue curves: The $T_{sqz}$-$\xi$ relation under different gripper forces. The yellow dash curve: The $T_{sqz}$-$\xi$ relation when $F_{grpr}$ is at its minimum value. (b) Black curves: The output torque-angle relation when a theoretically optimal spring is used ($\xi$=$19.27 N\cdot mm/^{\circ}$, $F_{grpr}$=125$N$). Red curves: The output torque-angle relation when a best commercially available spring is used ($\xi$=$6.00 N\cdot mm/^{\circ}$, $F_{grpr}$=125$N$).}
\label{output_tk}
\end{center}
\end{figure}

The relations between $T_{sqz}$, $T_{stch}$, and $T_{sprg}$ are visualized in Fig.\ref{output_tk}(a). Here, $\alpha$ is set to a constant $ 45^{\circ}$ and $d_{fgr}$ is set to a constant $45mm$ to reduce the number of variables. The vertical axis indicates the output values of $T_{sqz}$ or $T_{stch}$. The horizontal axis indicates the coefficient of the torsional springs, $\xi$. The solid blue curves show the $T_{sqz}$-$\xi$ relation under different $F_{grpr}$ values. The yellow dash curve shows the $T_{stch}$-$\xi$ relation ($F_{grpr}$ is passive in computing the yellow dash curve). The graph shows that $T_{stch}$ increases as the coefficient of the torsional springs increases. At the same time, the gripper will need a larger force to produce the same amount of $T_{sqz}$.

Considering this trade-off, we propose to select a proper $\xi$ by optimizing equation \eqref{output_minimize}:
\begin{equation}
    argmax_\xi\int ^{\alpha_{init}}_{\alpha_{end}}\left | T_{sqz}(\alpha, \xi)\cdot T_{stch}(\alpha, \xi)\right |d\alpha
\label{output_minimize}
\end{equation}
The equation computes the definite integral of $| T_{sqz}(\alpha, \xi)\cdot T_{stch}(\alpha, \xi) |$ under bounds $\alpha_{init}$ and $\alpha_{end}$. $\alpha_{init}$ and $\alpha_{end}$ are previous mentioned in Fig.\ref{output-direction}. They are the angles at the stretching extreme and the squeezing extreme, repectively. The reason we propose this optimization is the $\xi$ that induces the maximum integrand will make $T_{sqz}$ and $T_{stch}$ simultaneously large.

The optimal $\xi_{balanced}$ by solving equation \eqref{output_minimize} is $19.52N\cdot mm/^{\circ}$. The black curves in Fig.\ref{output_tk}(b) show the changes of $T_{sqz}$ and $T_{stch}$ under this value with repsect to a varying $\alpha$. The solid black curve is $T_{sqz}$($\alpha$). The dashed black curve is $T_{stch}$($\alpha$). From their positions, we get that under $\xi_{balanced}=19.52N\cdot mm/^{\circ}$, $T_{sqz}$ is close to $T_{stch}$ at all $\alpha$ angles, which makes the tool rotate smoothly and steadily in a whole squeezing-stretching cycle.

Unfortunately, commercial torsional springs with the theoretically optimal $\xi_{balanced}$ are too large to fit the size of our tool. Thus, we give up this optimal value and choose a spring with $\xi = 6.00 N\cdot mm/^{\circ}$ instead. This spring's coefficient value is most near to the optimal one. Meanwhile, it meets the dimensional requirements. The performance of the chosen spring is shown by the red curves in Fig.\ref{output_tk}(b). Like the black curves, the solid red one shows the $T_{sqz}$-$\alpha$ relation. The dashed curve shows the $T_{stch}$-$\alpha$ relation.

Under this spring selection, the designed tool can provide forces to fasten the screw sizes shown in the grey area of Table.\ref{table-1}. Here, a Robotiq Hand-E gripper with a maximum $125N$ grabbing force is assumed to squeeze and stretch the tool. The values in the table are from the Japanese Industrial Standards for general machinery (JIS B \cite{jis20091082} \cite{jis20081083}). The tool can output torque between 3.83$N\cdot m$$\sim$4.04$N\cdot m$ in the squeezing phase and between 0.66$N\cdot m$$\sim$0.85$N\cdot m$ in the stretching phase. According to the table, it could maximumly fasten M5 screws with 4.8 property class. Note that since the stretching force is weak, the tool may get stuck after a strong squeeze. To avoid this problem, one may choose to use a single-ratchet mechanism instead of the double-ratchet one. A single-ratchet tool only produces rotation in the squeezing phase. The stretching phase is free of load. The maximum fastening force of a single-ratchet tool equals to the maximum squeezing force.
\begin{table}[!htbp]
\caption{Fastenable Screw Sizes}
\label{table_example}
\begin{center}
\begin{tabular}{|c|c|c|c|c|c|}
\hline
\multirow{3}{*}{Screw size}&\multicolumn{5}{c|}{Tightening torque ($N\cdot m$)}\\
\cline{2-6}
&\multicolumn{5}{c|}{Property class}\\
\cline{2-6}
&4.8&6.8&8.8&10.9&12.9\\
\rowcolor{black!20}
\hline{}
\cellcolor{white}M3&0.56&1.10&1.45&2.08&2.43\\
\hline{}
M3.5&\cellcolor{black!20}0.89&\cellcolor{black!20}1.73&\cellcolor{black!20}2.28&\cellcolor{black!20}3.27&\cellcolor{black!20}3.82\\
\hline{}
M4&\cellcolor{black!20}1.31&\cellcolor{black!20}2.57&\cellcolor{black!20}3.38&4.84&5.66\\
\hline{}
M5&\cellcolor{black!20}2.65&5.19&6.80&9.78&11.43\\
\hline
M6&4.50&8.81&11.60&16.60&19.40\\
\hline
\end{tabular}
\end{center}
\label{table-1}
\end{table}

\subsection{Optimizing the Geometric Dimensions}
In this part, we optimize the rotational travel of the tool's output end as well as the geometric dimensions of the tool.

\subsubsection{The maximum rotational travel}
\begin{figure}[!htbp]
    \begin{center}
    \includegraphics[width=0.48\textwidth]{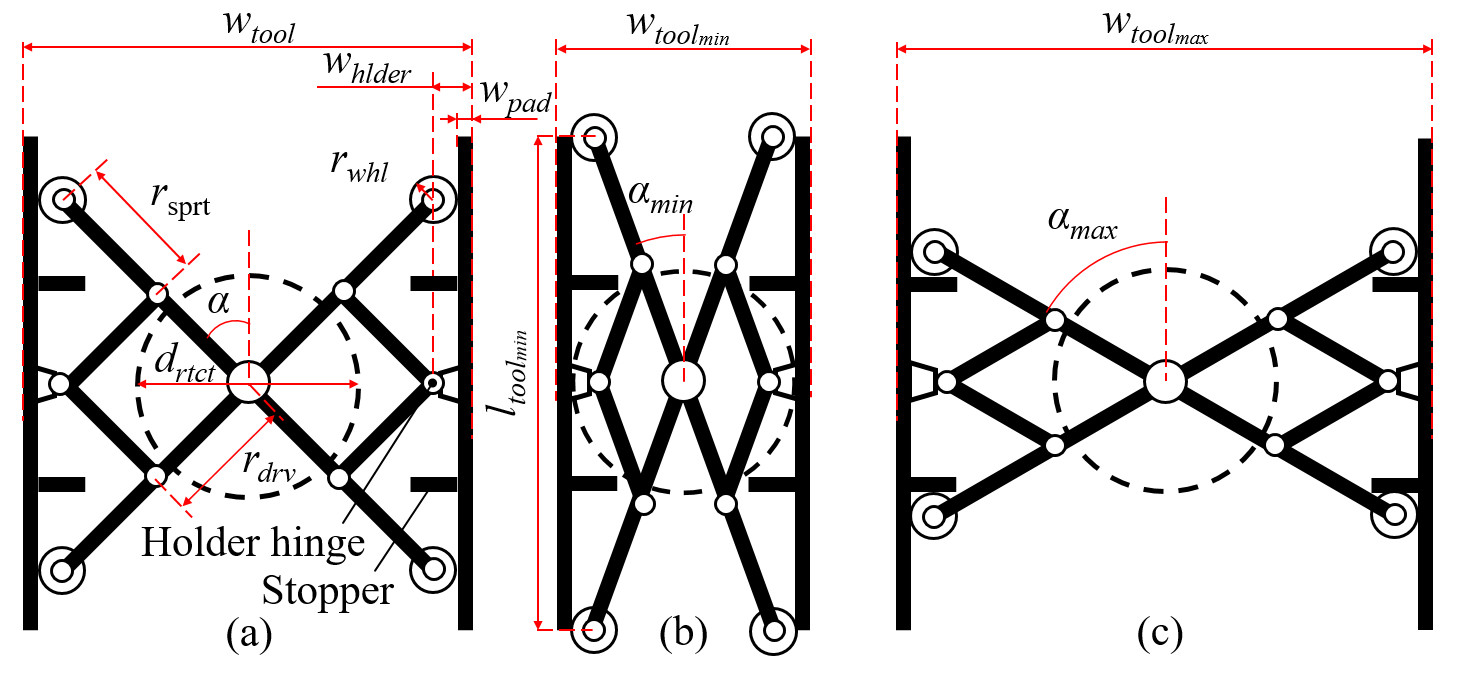}
    \caption{The various parameters studied in the optimization. (a) The parameters at an intermediate state. (b) The parameters at an end (closed) state. (c) The parameters at an initial (open) state.}
    \label{fbd-extreme}
    \end{center}
\end{figure}

Our geometric optimization's top priority is to maximize the rotational travel of the tool's output end. As shown in Fig.\ref{fbd-extreme}, the difference between $\alpha_{max}$ and $\alpha_{min}$ represents the maximum rotation travel, which equals to
\begin{equation}
\Delta\alpha_{max}=\sin^{-1}{\frac{w_{tool_{max}}-2r_{whl}}{4r_{drv}}}-\sin^{-1}{\frac{w_{tool_{min}}-2r_{whl}}{4r_{drv}}}.
\end{equation}
The value is dependent on four parameters $r_{drv}$, $w_{tool_{max}}$, $w_{tool_{min}}$, and $r_{whl}$. The constraints induced by each of them are as follows.

(i) $r_{drv}$. $\Delta\alpha_{max}$ monotonically increases with $1/r_{drv}$, thus $r_{drv}$ should be small. On the other hand, $r_{drv}$ should be designed to meet the requirements of the maximum output torque, as shown in equations \eqref{output_sqz} and \eqref{output_stch}. 

(ii) $w_{tool_{max}}$ and $w_{tool_{min}}$. Their values must meet the following equation 
\begin{equation}
w_{tool}=4r\sin{\alpha}+2w_{hldr},
\label{opimization-1}
\end{equation}
where $w_{hldr}$ is the distance from the outer surface of a holding pad to its hinge center. The equation shows that $w_{tool}$ monotonically increases with $\alpha$. In an initial state shown in Fig.\ref{fbd-extreme}(c), the supporting wheels are halted by two stoppers and $\alpha$ reaches maximum $\alpha_{max}$. The width of the tool reaches maximum $w_{tool_{max}}$. This value shall be smaller than the jaw width of a robotic gripper. In a final state shown in Fig.\ref{fbd-extreme}(b), when the holding pads contact the ratchet units, $\alpha$ reaches $\alpha_{min}$ and the width of the tool reaches $w_{tool_{min}}$. This value should be larger than the diameter of the ratchet.

(iii) $r_{whl}$. This value shall be as small as possible. It is limited by the availability of commercial products.

Considering these constraints, we determine the ratchet first. Then, considering the diameter of the ratchet and the jaw width of an expected robot gripper, we decide $w_{tool_{min}}$ and $w_{tool_{max}}$. Finally, we optimize $r_{drv}$ following the selected spring (Section.IV.B) and the torque constraints discussed in (Section.IV.A.3).

\subsubsection{The minimum height}

The second goal of our geometric optimization is to reduce the height of the tool. The distance between the two supporting wheels reaches a maximum value at a final state shown in Fig.\ref{fbd-extreme}(b). The lengths of the holding pads must be equal to or longer than this value. Thus, we name this value $l_{tool_{min}}$. $l_{tool_{min}}$ is computed as
\begin{equation}
l_{tool_{min}}=2(r_{drv}+r_{sprt})\cos{\alpha_{max}}.
\label{opimization2}
\end{equation}
This equation can be easily obtained from the geometric relations in Fig.\ref{fbd-extreme}(b). The equation indicates that the length of the holding pad is affected by three parameters $r_{drv}$, $r_{sprt}$, and $\alpha_{max}$. Since $r_{drv}$ and $\alpha_{max}$ have been determined in the previous optimization, we focus on $r_{sprt}$, the length of the supporting arm, and study how to reduce it. 

\begin{figure}[!htbp]
\begin{center}
\includegraphics[width=0.48\textwidth]{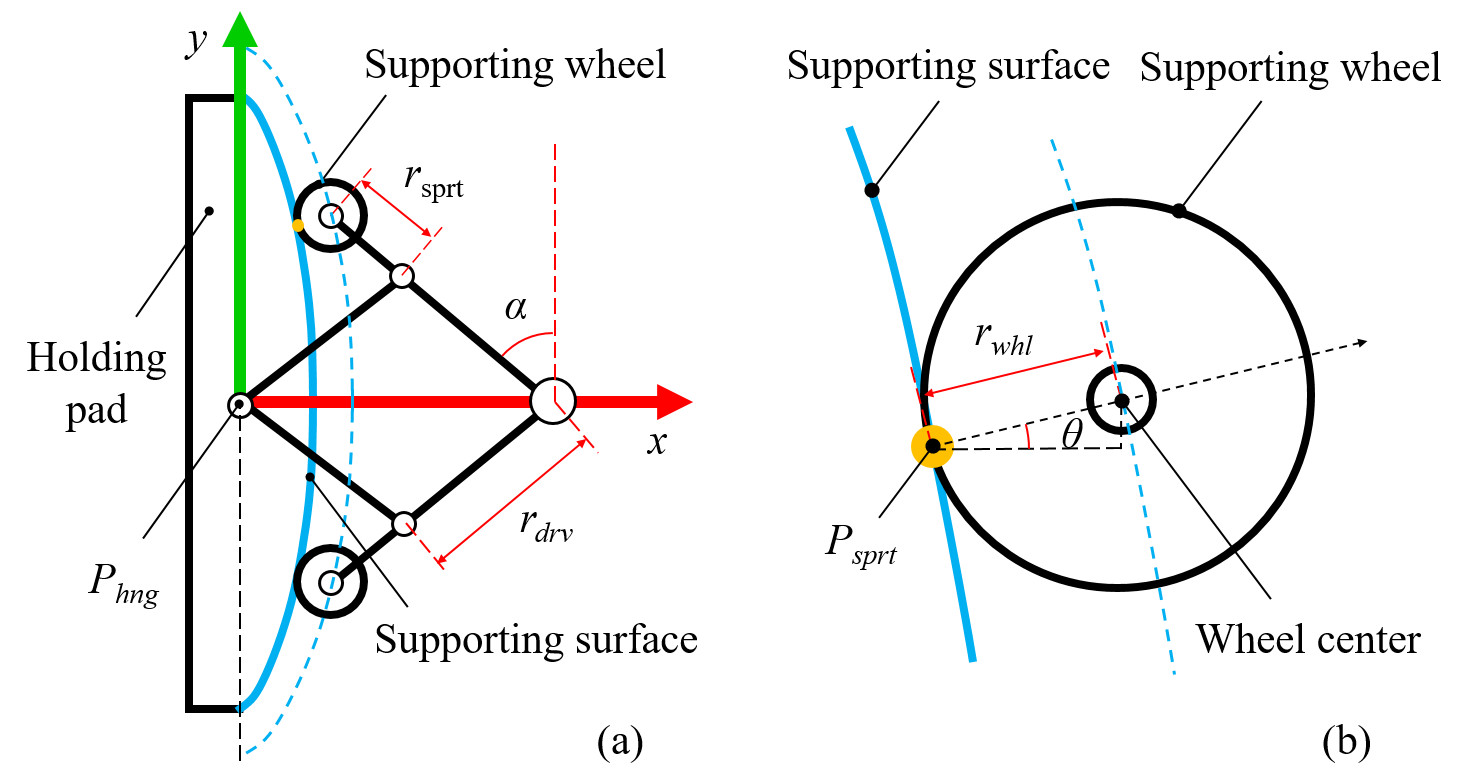}
\caption{(a) Introducing a curved supporting profile to the inner surface of a holding pad. The solid blue curve is the supporting profile. The blue dash curve is the trajectory of the supporting wheel center. (b) A close-up view of the supporting wheel.}
\label{fbd-sptwheel}
\end{center}
\end{figure}

\paragraph{Curved supporting surface}
The idea we use to lessen $r_{sprt}$ is to introduce a curved profile to the holding pads' inner surface. The curved profile converts the wheel and the holding pad pair into a mechanical cam pair. The blue curve in Fig.\ref{fbd-sptwheel} exemplifies such a curved profile. The design shortens $r_{sprt}$ and thus compresses the lengths of the holding pads.

We can derive the parametric form of the curved profile as follows. We set up a reference frame at $P_{hnge}$. The $x$ axis of the frame points to the center of the ratchet. The $y$ axis aligns with the holding pad. They are illustrated by the red and green arrows in Fig.\ref{fbd-sptwheel}. Suppose the center of the supporting wheel in the frame is ($x_{whl}$, $y_{whl}$), we can get the following relations:
\begin{equation}
    \alpha=\sin^{-1}{\frac{x_{whl}}{r_{drv}-r_{sprt}}},
\end{equation}
\begin{equation}
    y_{whl}=(r+r_{sprt})\cos{\alpha}.
\end{equation}
The curve implied by $f_{whl}$=($x_{whl}$, $y_{whl}$) is the trajectory of the supporting wheel center. The blue dash line in Fig.\ref{fbd-sptwheel}(b) shows this curve. The profile of the holding pad's inner surface is essentially the trajectory of the contact point $P_{sprt}$. It can be obtained by shifting the curve $f_{whl}$ with an offset $r_{whl}$ along its reversed normal direction. Since the reversed normal direction equals to:
\begin{equation}
    \theta(\alpha)=\tan^{-1}{\frac{-1}{f^{'}_{whl}}},
\end{equation}
the profile of the holding pad's inner surface consequently has the following parametric form:
\begin{equation}
    \left\{\begin{matrix}
    \begin{aligned}
        &x_{sprt}=x_{whl}+r_{whl}\cos{\theta(\alpha)} \\
        &y_{sprt}=y_{whl}(\alpha)-r_{whl}\sin{\theta(\alpha)} 
    \end{aligned}
    \end{matrix}\right..
    \label{curvepara}
\end{equation}
Here, ($x_{sprt}$, $y_{sprt}$) indicates the supporting point under the reference frame. $x\in (0, r-r_{sprt})$. Equation \eqref{curvepara} is the parametric form of the curved profile. Its shape can keep the contact between the finger pad and the supporting wheel along with the rotation process. Meanwhile, it forces the motion of the SLE arms to a rotation around the ratchet center.

\paragraph{Structural stability}
The disadvantage of a curved profile is that it decreases the structural stability of the tool. This is a trade-off caused by reducing the length of $r_{sprt}$. A smaller $r_{sprt}$ lessens $l_{tool_{min}}$ but makes the structure less stable. Thus, we perform an optimization for $r_{sprt}$, and formulate the problem as finding a smallest $r_{sprt}$ that has satisfying stability.

\begin{figure}[!htbp]
\begin{center}
\includegraphics[width=0.48\textwidth]{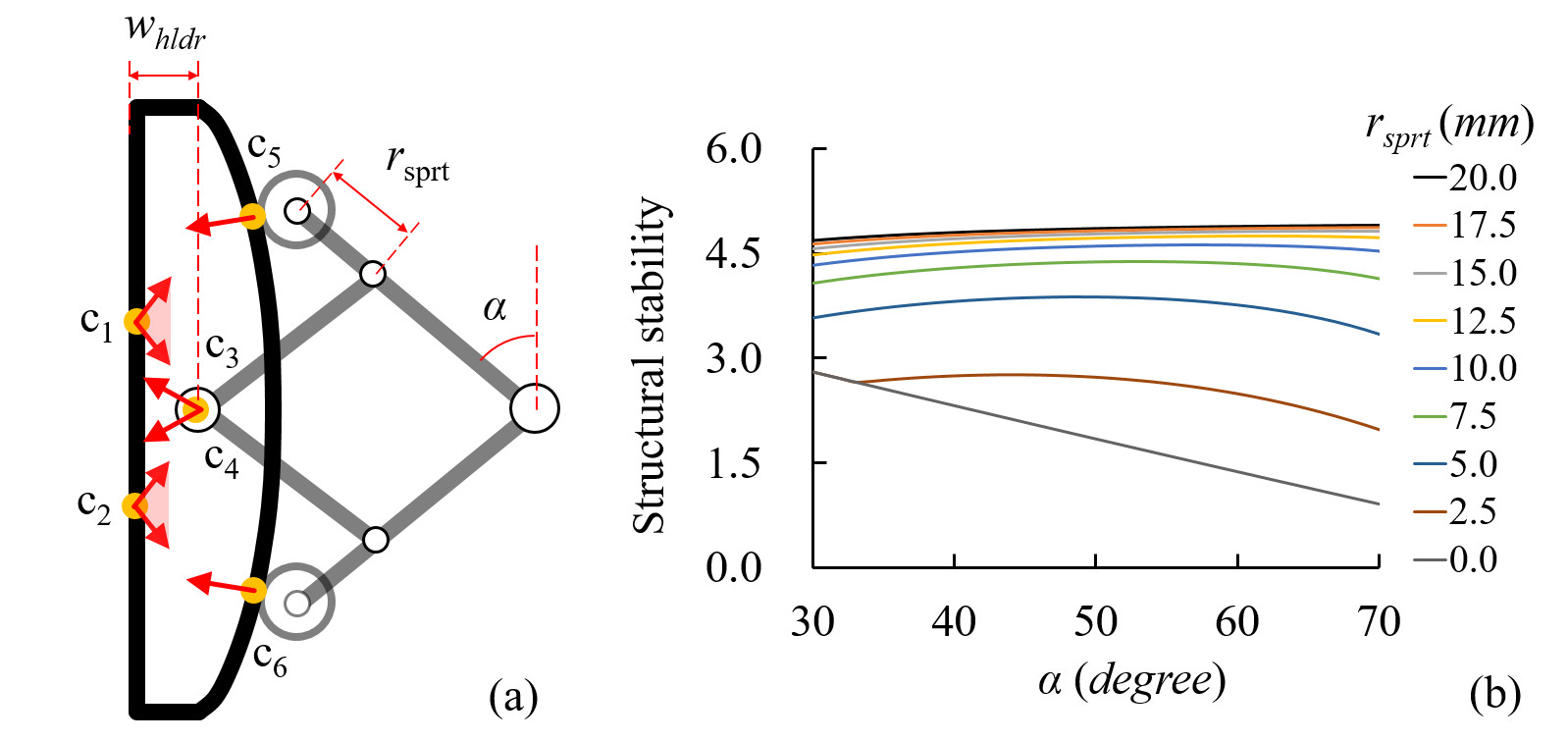}
\caption{(a) We formulate the evaluation of the structural stability into measuring the wrench cone formed by the grasp wrench set at six contact points $c_{1}\sim c_{6}$. (b) The changes in the structural stability with respect varying $r_{sprt}$ and $\alpha$. The horizontal axis is the $\alpha$ angle. The curves with different colors are the results of different $r_{sprt}$ values. The $w_{hldr}$ used to get the results is set to $6.5mm$.}
\label{stabilityindex}
\end{center}
\end{figure}

We evaluate the stability by measuring the wrench cone formed by the Grasp Wrench Set (GWS) of contact points on the holding pads. To simplify the model, we assume the motion is 2D, and the contact surfaces are rigid and smooth. Fig.\ref{stabilityindex}(a) shows all the contact points on the holding pad. We represent them using symbols $c_{i} (i=1\sim  6)$, where $c_{1}$ and $c_{2}$ denote the contacts with by the robot gripper. $c_{3}$ and $c_{4}$ denote the contacts with the supporting wheel. The connections at the hinge are represented as two contacts $c_{5}$ and $c_{6}$. At each $c_{i}$, we use $f_{i}$ and $\tau_{i}$ to denote the exerted force and moment. We assume that there are friction forces at $c_{1}$ and $c_{2}$ but $c_{3}\sim c_{6}$ are frictionless, as $c_{3}\sim c_{6}$ are on the surface of supporting wheels or are hinges. The wrench $\omega$ exerted by all the contacts thus equals to:
\begin{equation}
    \omega=\sum_{i=1}^{6}G_{i}
\begin{bmatrix}
\bm{f_{i}}\\ 
\tau_{ni}
\end{bmatrix},
\end{equation}
\begin{equation}
    G_{i}=
\begin{bmatrix}
\bm{I}&0\\ 
\left[\bm{p_{i}\times}\right]&\bm{n_{i}}
\end{bmatrix},
\end{equation}
where $p_i$ and $n_{i}$ indicate the contact positions and the contact normals.

Then, we use the Minkovski sum of the wrenches to find the wrench cone following \cite{harada2014stability}. The structural stability index ($Q$) is computed as the minimum distance from the wrench cone boundary to the origin wrench space \cite{ferrari1992planning}. Fig.\ref{stabilityindex}(b) shows the results of our computation under varying $r_{sprt}$ and $\alpha$. The results imply that the structural stability decreases with the increase of both $\alpha$ and $r_{sprt}$. The decreasing speed significantly rises when $r_{sprt}$ is shorter than 10.0$mm$. Thus, we fabricate the tool by using $r_{sprt}$=10.0$mm$.

\section{PROTOTYPING THE DESIGN}
\begin{figure}[!htbp]
\begin{center}
\includegraphics[width=0.48\textwidth]{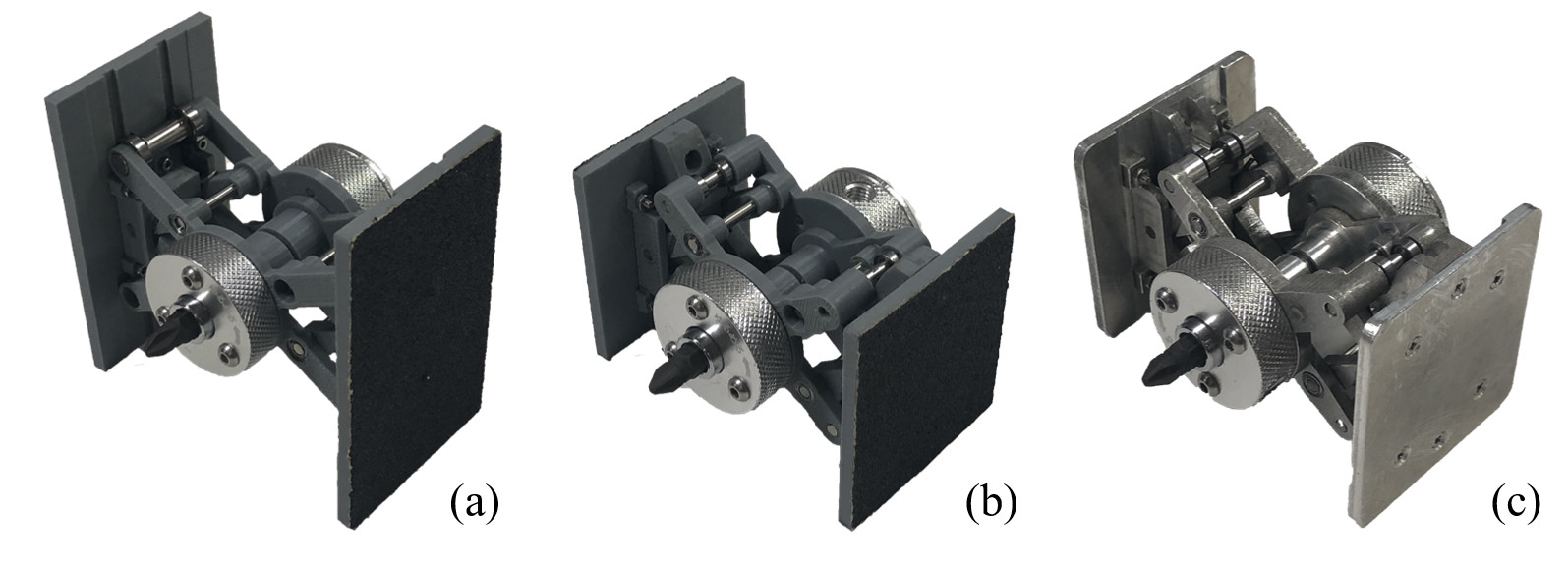}
\caption{Three versions of prototypes. (a) The version with flat holding pads. (b) The version with curved holding pads. (c) The aluminum version of the second prototype. It is used in the real-world applications developed in the experimental section.}
\label{threeprototypes}
\end{center}
\end{figure}

Following the optimization discussed in the previous section, we fabricated three versions of prototypes. Especially, we compare two major versions. The first major version is the one with flat holding pads, as shown in Fig.\ref{threeprototypes}(a). The second major version uses holding pads with curved profiles, as shown in Fig.\ref{threeprototypes}(b). We compare them from the perspectives of dimension, output torque, and output angle. The frames of these two versions are made by using a 3D printer (ABS materials). Besides them, we further made an aluminum version of the second prototype, as is shown in Fig.\ref{threeprototypes}(c). This version has better structural strength and is used to develop real-world experiments in the experimental section.

\begin{figure}[!htbp]
\begin{center}
\includegraphics[width=0.48\textwidth]{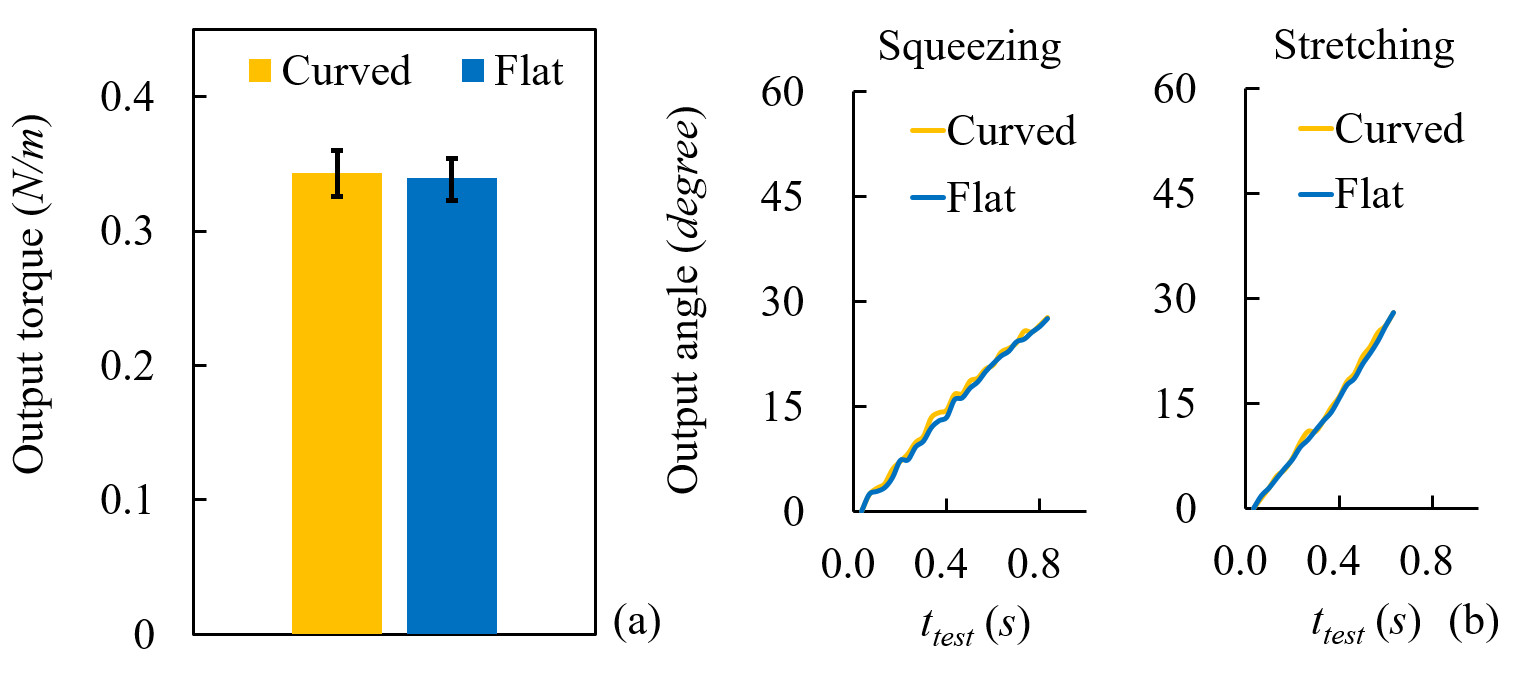}
\caption{Comparing the output performance of the first and second prototypes. The results of the prototype with curved holding pads are shown in yellow. The results of the prototype with flat holding pads are shown in blue. (a) The maximum output torque under a 20$N$ gripping force. The results are based on 5$times$ of repetition. (b) The changes in rotation angles under a 55$mm/s$ grasp velocity.}
\label{twoprototypecomparison}
\end{center}
\end{figure}

Note that the images in Fig.\ref{threeprototypes} are taken on the same scale. By listing them side-by-side, we can observe that the curved profile (Fig.\ref{threeprototypes}(b)) significantly reduces the dimension of the tool made by flat holding pads (Fig.\ref{threeprototypes}(a)). The widths of the two prototypes are both 70$mm$. The height of the holding pad decreases from 80$mm$ to 54$mm$, leading to a more compact design. Meanwhile, although the dimension is significantly reduced, the second major version has similar output performance as the first one. Fig.\ref{twoprototypecomparison} shows the details. Here, the maximum output torques of both the two major prototypes are tested with a 20$N$ gripping force. The rotation angles are tested with a 40$mm/s$ grasp velocity in the squeezing phase, and a 55$mm/s$ grasp velocity in the stretching phase. The results indicate that there is no significant difference between the output performance of the two prototypes.

Table.\ref{table_para} lists up the various parameters we used to fabricate the first and second versions of prototypes. The ratchet units used to make the double-ratchet mechanism are widely available commercial products\footnote{ANEX, http://www.anextool.co.jp/item/316}. It finalizes the diameter of the ratchet (32$mm$). The remaining parameters are the optimal values found considering the constraints discussed in the previous section.
\begin{table}[!htbp]
\begin{center}
\centering
\setlength{\tabcolsep}{0.1mm}{
 \caption{The Parameters Used to Prototype the Tool}
 \label{table_para}
 \begin{tabular}{lll}
  \toprule
  Parameter~ & Value & Definition \\
  \midrule
  $w_{tool_{max}}$  & $83mm$  & 
  Maximum width between two holding pads \\
  $w_{tool_{min}}$  & $40mm$  & 
  Minimum width between two holding pads  \\
  $w_{hldr}$   & $6.5mm$ &
  Dist. from a holding surface to the hinge center\\
  $w_{pad}$      & $2mm$   &
  Thickness of the holding pads \\
  $r_{drv}$    & $20mm$  &
  Length of the driving arm (mC-SLE)\\
  $r_{sprt}$ ($ver.1$)~~ & $20mm$  &
  Length of the supporting arm (mC-SLE)\\
  $r_{sprt}$ ($ver.2$)~~ & $10mm$  &
  -\\
  $r_{whl}$    & $1.5mm$   &
  Radius of the supporting wheel (mC-SLE)\\
  $l_{tool}$ ($ver.1$) & $80mm$ & 
  Height of the holding pads \\
  $l_{tool}$ ($ver.2$) & $54mm$ & 
  - \\
  $\xi$          & $6.00N\cdot mm/^{\circ}$~~ &
  Elastic coefficient of the torsional spring\\
  $d_{rtct}$          & $32mm$&
  Diameter of the ratchet\\
  \bottomrule
 \end{tabular}}
\end{center}
\end{table}

\section{THE MANIPULATION POLICIES TO USE THE TOOL}
This section develops the manipulation policies for using the designed tool. It consists of three parts: (1) Recognizing the tool; (2) Automatically planning grasps and manipulation sequences; (3) Exchanging the tooltips.

\subsection{Recognizing the Tool}
\begin{figure}[htbp]
\begin{center}
\includegraphics[width=0.48\textwidth]{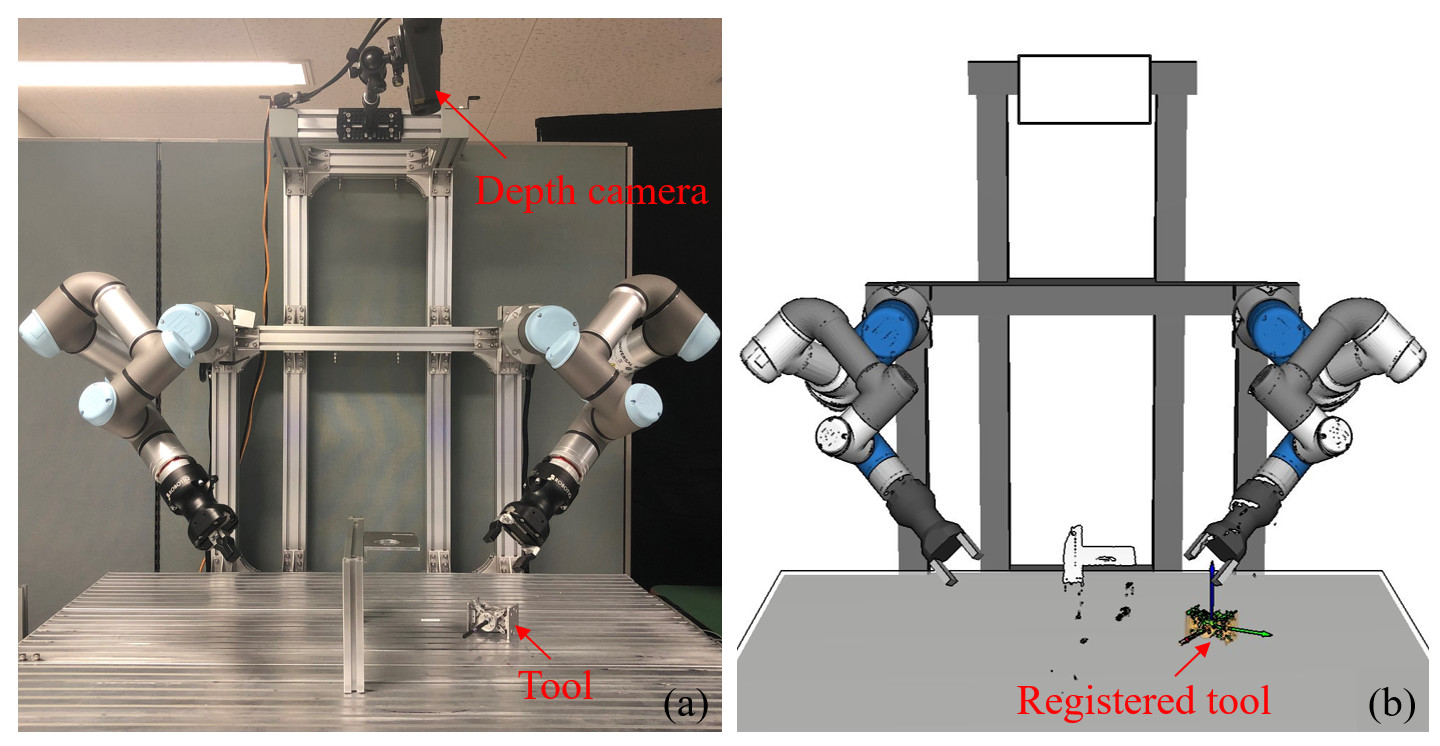}
\caption{Recognizing the tool using a depth sensor. (a) The real-world system setting. (b) The matched tool model in the simulation environment.}
\label{recognizing}
\end{center}
\end{figure}

We assume a depth sensor to be employed for visual recognition. One may use a depth sensor to scan the workspace and locate the tool by registering its model to the collected point cloud. Conventional algorithms like DBSCAN-based segmentation \cite{witten2002data}, RANSAC-based global search \cite{zhou2016fast}, and ICP-based local refinement \cite{pomerleau2013comparing} are employed in the registration. Fig.\ref{recognizing} shows an example of the collected point cloud and the matched tool pose using the mentioned rough estimation and local refinement routine.

\subsection{Picking-up and Manipulation}
We use the grasp planning methods presented in \cite{harada2014stability}\cite{tsuji2014grasp} to plan grasping poses, and use the methods presented in \cite{wan2016integrated}\cite{wan2017teaching}\cite{wan2019preparatory}\cite{wan2016achieving} to plan placements and regrasp sequences. 

\begin{figure}[!htbp]
\begin{center}
\includegraphics[width=0.48\textwidth]{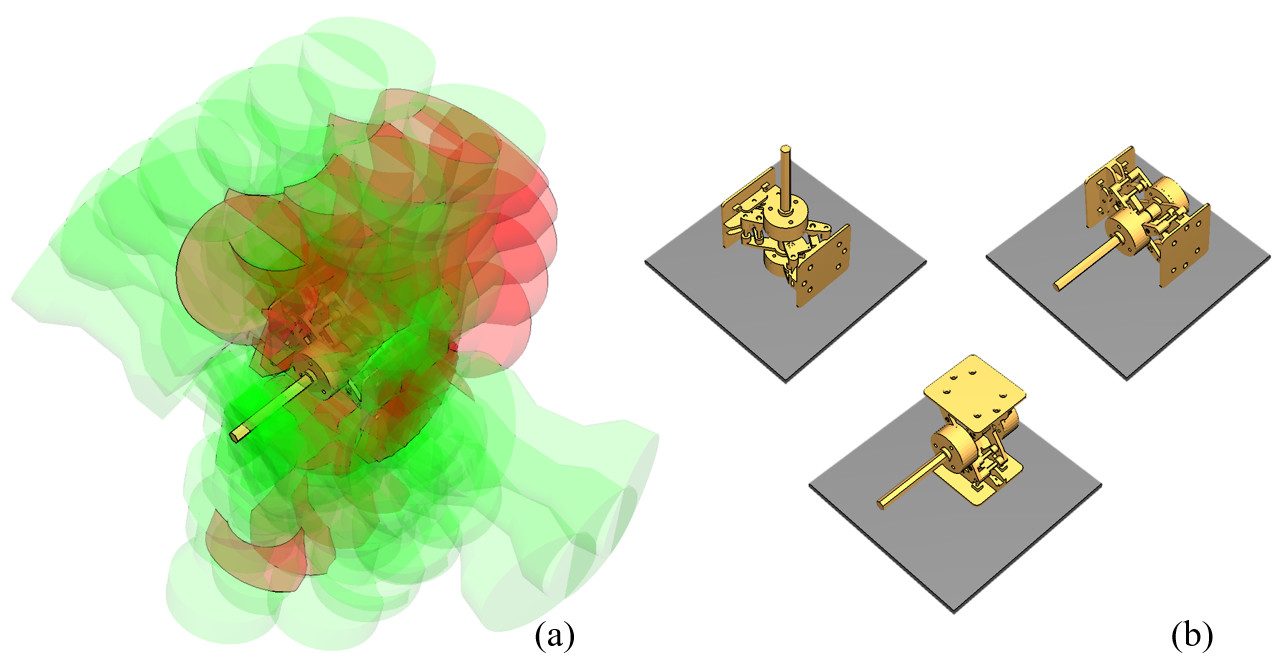}
\caption{(a) Grasp candidates without considering the environmental constraints. The red hand indicates the tool-control grasp. (b) Stable placements of the tool.}
\label{motionplanning-grasp}
\end{center}
\end{figure}

Fig.\ref{motionplanning-grasp}(a) exemplifies some grasp candidates found by a grasp planner. The plan is automatically performed using the mesh model of the tool. The red hands refer to the grasp poses that can use the tool. These poses are named as the tool-control grasp poses. The green hands refer to the grasp poses that can only hold the tool. They are called the tool-holding grasp poses. 

Fig.\ref{motionplanning-grasp}(b) shows the stable placements of the tool on a table. The tool could be placed at an arbitrary position on a worktable with any of these stable placements.

\begin{figure}[!htbp]
\begin{center}
\includegraphics[width=0.48\textwidth]{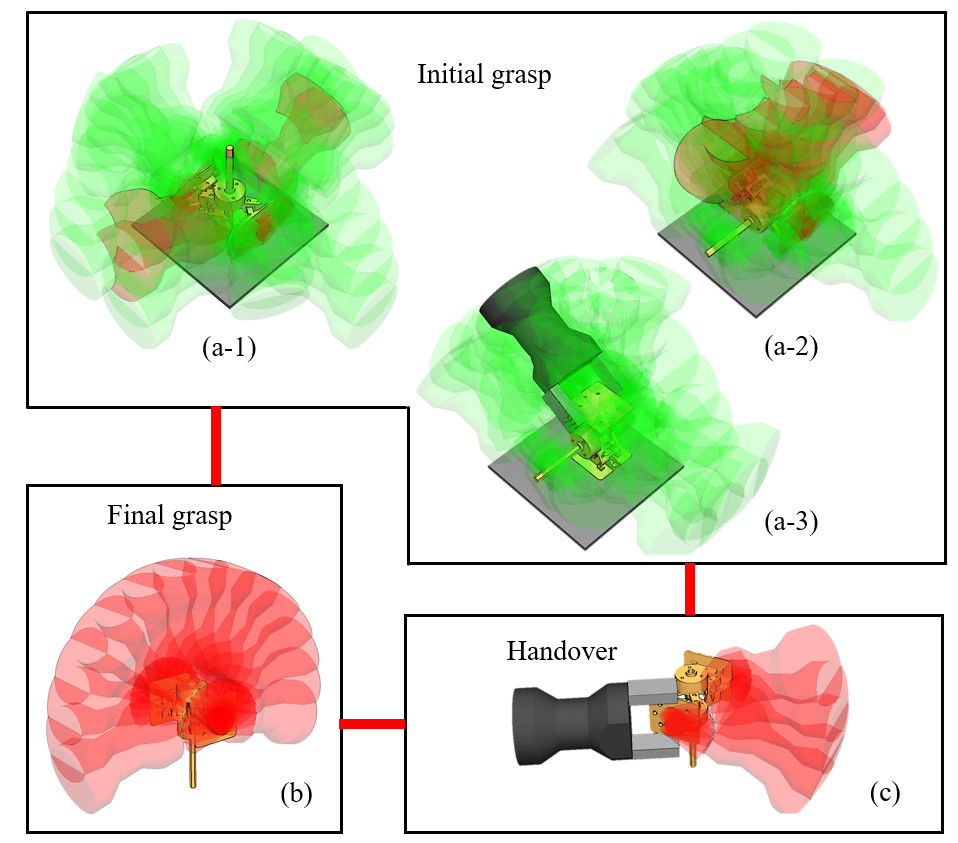}
\caption{(a) Initial grasp poses. They are the collision-free grasp candidates for the tool placed on a table. (b) Final grasp poses. (c) The grasp poses for handover.}
\label{motionplanning-regrasp}
\end{center}
\end{figure}

Fig.\ref{motionplanning-regrasp}(a) further shows the collision-free tool-control and tool-holding grasp poses for each of the stable placements. Given a specific task, the goal position of the tool can be generated according to the position of the objective screw. Then, based on the goal position, the collision-free tool-control grasp poses of the tool at the goal can be generated, as shown in Fig\ref{motionplanning-regrasp}(b). To compute the tool-control grasp poses, besides the collision constraints, the geometric backtracking between the tool's initial pose and the final goal pose shall also be considered. In the case that there is no shared tool-control grasp poses between the initial tool pose and the goal tool pose, the planner will trigger a handover to change the tool pose. For example, there is no tool-control pose in the initial grasp poses shown in Fig.\ref{motionplanning-regrasp}(a-3). Thus, a handover regrasp is necessary, as shown in Fig.\ref{motionplanning-regrasp}(c).

\subsection{Exchanging the Tooltips}
The tool has two hexagonal magnetic sockets at its output ends for attaching tooltips. A group of tooltips with hexagonal connecting ends can be exchanged and attached to the sockets to meet various task requirements. Attaching the tooltip into the socket is essentially a peg-in-hole task. A compliant strategy integrating linear, spiral, and rotation search with impedance control is developed to conduct this task.
\subsubsection{Linear search}
\begin{figure}[htbp]
\begin{center}
\includegraphics[width=0.48\textwidth]{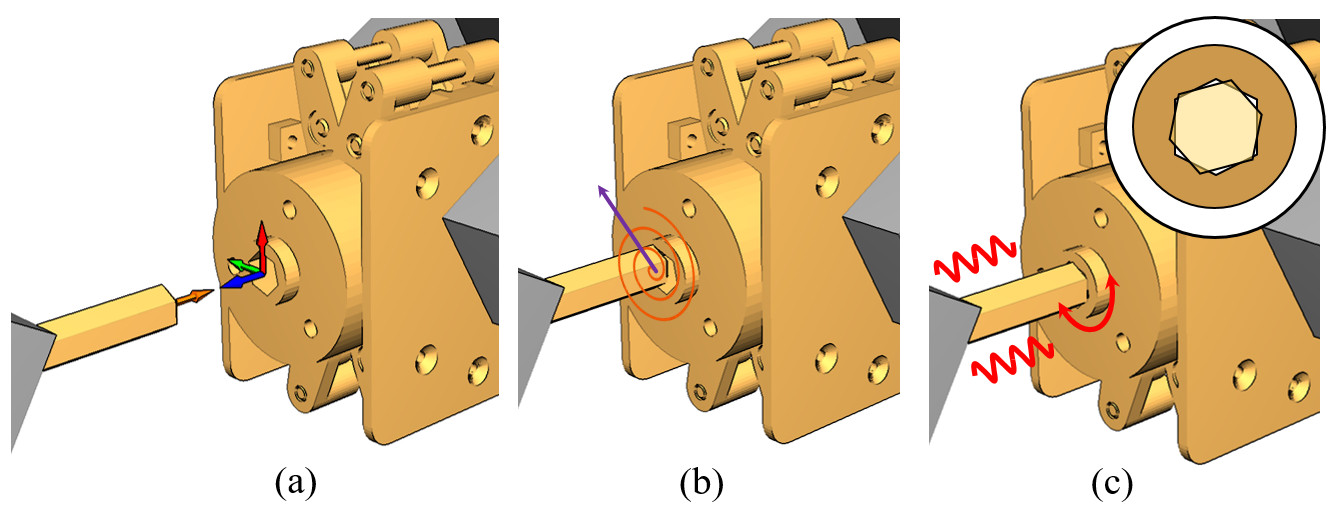}
\caption{(a) Linear search. The hand holding the tooltip moves linearly along the orange vector until it hits the connecting surface. (b) Spiral search. The orange spiral curve indicates the generated spiral path. The purple vector shows the initial spiral direction. (c) Rotation search and impedance control.}
\label{forcecontrol}
\end{center}
\end{figure}

In linear search, the robotic hand holding the tooltip moves along a straight line to make the tooltip end contact the pre-insertion surface. An example is shown in Fig.\ref{forcecontrol}(a). The robot hand in the example moves along an orange direction $\mathbf{v}^{att}$ to search for the contact between the tooltip end and the socket. The linear motion stops when equation \eqref{linear-rsch} is satisfied.
\begin{equation}
    \left | \mathbf{v}^{att}(\mathbf{R}^{grpr} \cdot \mathbf{F}^{grpr})\right | \geq F_{threshold}.
\label{linear-rsch}
\end{equation}
Here, $\mathbf{R}^{grpr}$ is the rotation matrix of the holding hand, $\mathbf{F}$ is the observed force from the F/T sensor mounted at the wrist. $F_{threshold}$ is the desired contact threshold.

\subsubsection{Spiral search}
Assume that the tooltip end stops at position $\mathbf{P}^{hnd}_0$ at the end of linear search, as is shown in Fig.\ref{forcecontrol}(b), then, based on this position, a spiral curve and spiral search is planned. The spiral curve is generated according to equation \eqref{spiral-rsch} in the $\mathbf{R}^{socket}_{x}$-$\mathbf{R}^{socket}_{y}$ plane. Here, $\mathbf{R}^{socket}$ represents the pose of the socket. The $x$ and $y$ at the subscript denote the local $x$ and $y$ axes of the rotation matrix.
\begin{equation}
  \mathbf{P}^{hnd}_{i+1}=r^{sprl}_{i+1}\cdot\mathtt{rodrigues}(\theta^{sprl}_{i+1}, \mathbf{v}^{att})\cdot\mathbf{v}^{sprl}+\mathbf{P}^{hnd}_{i}.
  \label{spiral-rsch}
\end{equation}
Here, $\mathbf{v}^{sprl}$ indicates the initial spiral direction (the purple vector in Fig.\ref{forcecontrol}(b)). $\mathbf{P}^{hnd}_{i}$ and $\mathbf{P}^{hnd}_{i+1}$ are the current position and the planned next position, respectively. $\mathtt{rodrigues}(\theta, \mathbf{v})$ is the Rodrigues' rotation formula. $\theta^{sprl}$ and $r^{sprl}_{i+1}$ are computed as as:
\begin{equation}
\theta^{sprl}_{i+1} = \theta^{sprl}_{i}+\delta\theta^{sprl},~r^{sprl}_{i+1} = r^{sprl}_{i}+\delta r^{sprl},
\end{equation}
where $\delta \theta^{sprl}$ and $\delta r^{sprl}$ are the discretized step rotation and step length of the spiral curve. Since the end of tooltip are chamfered, when the tooltip end is aligned to the pre-inserting position as shown in Fig.\ref{forcecontrol}(c), equation \eqref{spiral-rsch} will be violated and the robot will stop the spiral research.

\subsubsection{Rotation research and impedance control}
After the spiral search, rotation search and impedance control are applied to complete the insertion. We define the impedance control in the workspace following the conventional impedance control law:
\begin{equation}
  \mathbf{F}^{insrt} + \mathbf{F}^{rsst}_i = m\cdot\mathbf{\ddot{P}}^{hnd}_i+c\cdot\mathbf{\dot{P}}^{hnd}_i+k\cdot(\mathbf{P}^{hnd}_i-\mathbf{P}^{hnd}_{i-1})
\end{equation}
where $m$, $c$, and $k$ are inertia of the held object, damping coefficient, and stiffness respectively.  $\mathbf{\ddot{P}}^{hnd}_i$, $\mathbf{\dot{P}}^{hnd}_i$, and $\mathbf{{P}}^{hnd}_i$ are the acceleration, velocity, and displacement of the holding hand. $\mathbf{F}^{insrt}$ is the desired insertion force which points to the same direction as $\mathbf{v}^{att}$. $\mathbf{F}^{rsst}_i$ is the external force of the environment.
The generated hand motion is thus:
\begin{equation}
  \mathbf{P}^{hnd}_{i+1} = \frac{\mathbf{F}^{insrt} + \mathbf{F}^{rsst}_i+m\frac{(2\mathbf{P}^{hnd}_{i}-\mathbf{P}^{hnd}_{i-1})}{dt^2}+c\frac{\mathbf{P}^{hnd}_{i-1}}{dt}+k\mathbf{P}^{hnd}_{i}}{\frac{m}{dt^2}+\frac{c}{dt}+k}.
\end{equation}

Along with the impedance control, the hand holding the tool will rotate around $\mathbf{v}^{att}$ to perform rotation research. Thus, $\mathbf{F}^{rsst}_i$ changes with the environment contact and varies with rotation and insertion. The two robots stop simultaneously when $\mathbf{F}^{insrt}$ equals to $(-\mathbf{F}^{rsst}_{i})_{\mathbf{v}^{att}}$, namely when the tooltip end contacts with the bottom of the socket and the insertion is successfully conducted.

\section{EXPERIMENTS AND ANALYSIS}
We perform experiments using two UR3 robotic arms, with a Robotiq Hand-E two-finger parallel gripper mounted at the tool center point of each arm. The vision sensor used in the experiments is a Photoneo PhoXi 3D Scanner M. The sensor is mounted above the workspace for visual recognition. The experimental section consists of two main parts. In the first part, we study the performance of the design. We test the maximum output torque with respect to the designed parameters and examine the relationship between the gripping velocity and the rotation velocity. In the second part, we study the manipulation policies of the tool. We implement programs for the robot to exchange the tooltips and develop real-world applications to verify the proposed strategies.

\subsection{The Performance of the Design}
\subsubsection{Torque at the tooltip}
\begin{figure}[!htbp]
\begin{center}
\includegraphics[width=0.48\textwidth]{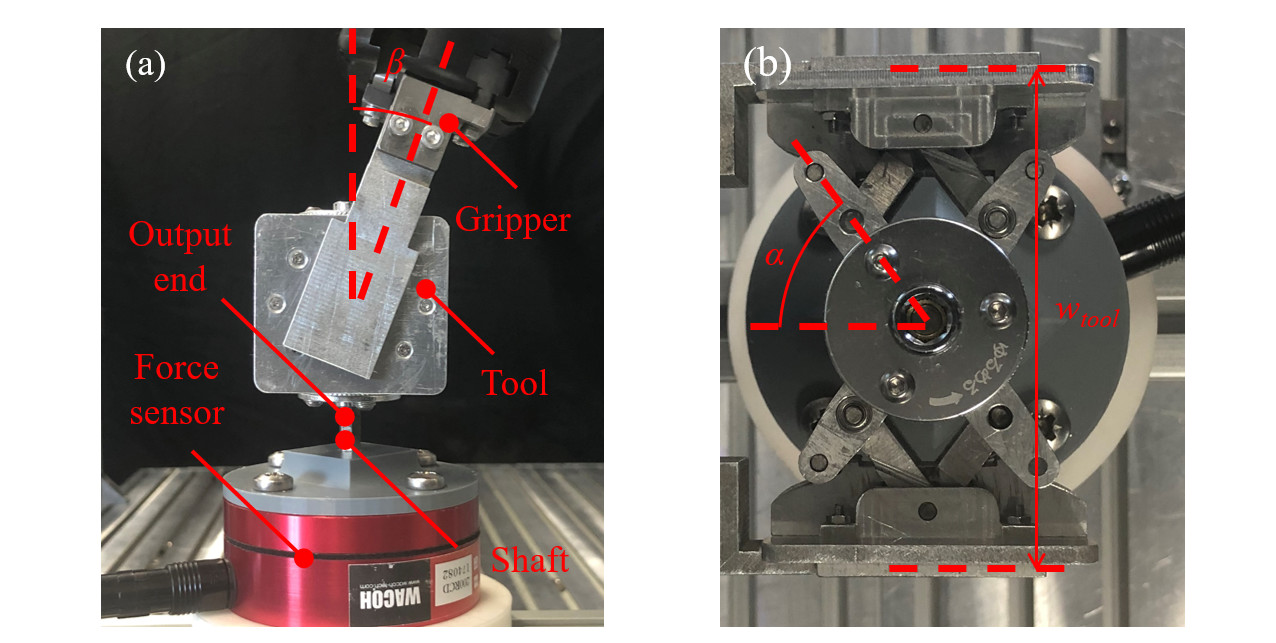}
\caption{The experimental setting for measuring the real maximum output torque. A DynPick Capacitive 6-axis force sensor (200N, WACOH-TECH Inc.) is used to measure the output torque. The tool is held by one Robotiq Hand-E gripper. The tooltip is inserted into a slot fixed on the torque sensor.}
\label{exp-output-set}
\end{center}
\end{figure}

This section presents the experiments used to measure the real output torque $T_{out}$ with respect to changing tool width $w_{tool}$ and holding angle $\beta$, respectively. The experimental setting is shown in Fig.\ref{exp-output-set}. A DynPick Capacitive 6-axis force sensor (200N, WACOH-TECH Inc.) is used to measure the torque values. One of the robotic grippers held the tool. The tooltip is inserted into a slot fixed on the torque sensor. The robotic gripper can close as well as open its jaw to exert force on the tool. The force sensor can thus measure the output torque on-line for both the squeezing and stretching phases. The stable peak torque measured by the force sensor is recorded as the maximum torque. 

\begin{figure}[!htbp]
\begin{center}
\includegraphics[width=0.48\textwidth]{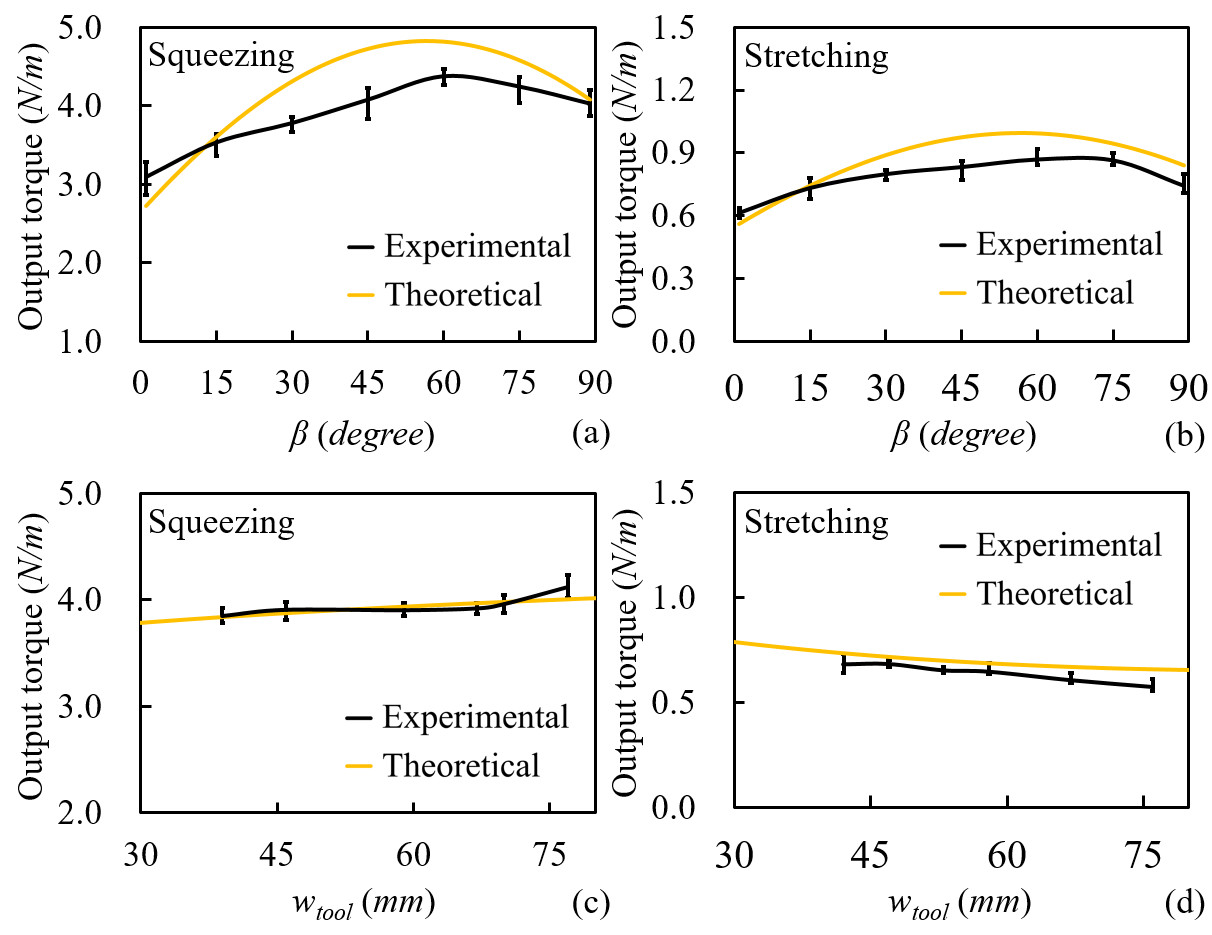}
\caption{The maximum output torque $T_{out}$ with respect to the tool width $w_{tool}$ and the holding angle $\beta$. Yellow curves indicate the theoretical values. Black curves indicate the experimental values. (a, b) The relation between the maximum output torque $T_{out}$ and a changing $\beta$. (c, d) The relation between $T_{out}$ nd a changing $w_{tool}$. The $\beta$ angle is selected to be a fixed value 90$^{\circ}$. The theoretical curves are computed using $\xi$=6$N\cdot mm/^{\circ}$ and $F_{grpr}$=125$N$.}
\label{exp-output-data}
\end{center}
\end{figure}

The results are shown in Fig.\ref{exp-output-data} where the yellow curves indicate the theoretical values, the black curves indicate the experimental values. The experimental results are identical to the theoretical analysis.

Fig.\ref{exp-output-data}(a, b) are the relation between the maximum output torque $T_{out}$ and a changing holding angle $\beta$. In both the squeezing and stretching phases, the maximum output torques increase with $\beta$ in the beginning, after it reaches the peak at $\beta$=60$^{\circ}$, the torque starts to decrease. The yellow curves and black curves have a similar tendency but a statistically observable difference. The difference is caused by the various errors like assembly errors, machining errors, imperfect rigid body assumption for the links, etc.

Fig.\ref{exp-output-data}(c, d) are the relation between $T_{out}$ and a changing $w_{tool}$. The $\beta$ angle is selected to be a fixed value 90$^{\circ}$. The maximum output torque slightly increases with $w_{tool}$ in the squeezing phase and slightly decreases in the stretching phase. The variations are minor, and the maximum output torques in both phases are nearly unchanged. Thus, one may ignore the influence of $w_{tool}$ in practice. Note that the theoretical curves are computed using $F_{grpr}$=125$N$ (maximum of Robotiq Hand-E) and $\xi$=6.00$N\cdot mm/^{\circ}$.

\subsubsection{Velocities at the tooltip}
\begin{figure}[htbp]
    \begin{center}
    \includegraphics[width=0.48\textwidth]{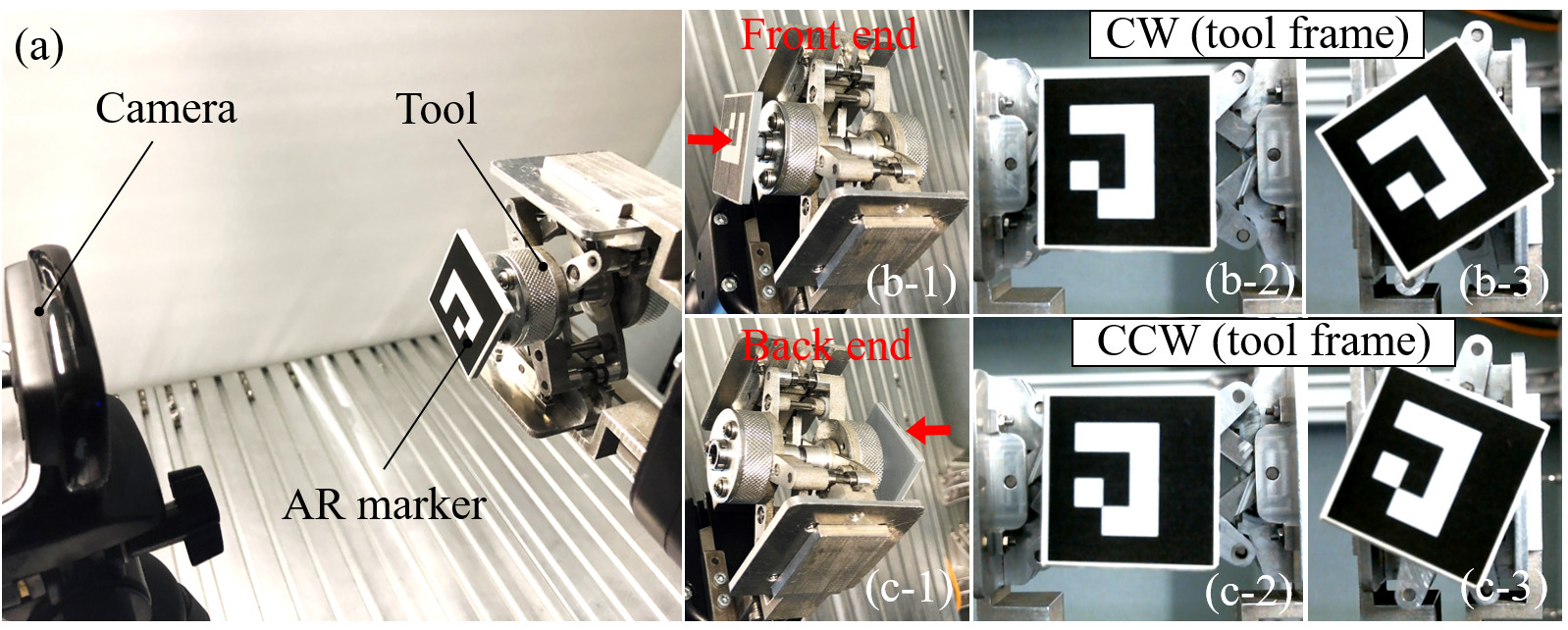}
    \caption{An AR marker is used to detect the rotating velocity. (a) The setting for tracking the rotation. (b) The clockwise rotation at the front end. (c) The counter-clockwise rotation at the back end. Note: The CW and CCW are counted in the tool frame.}
    \label{visionsetting}
    \end{center}
\end{figure}

This section presents the experiments used to examine the rotation of the tooltip. Particularly, we compared four cases. They are (1) the rotation in the squeezing phase, (2) the rotation in the stretching phase, (3) the rotation in a whole squeezing-stretching cycle, and (4) the rotation during a continuous rotation. An AR marker is used to assist in tracking the rotation. It is attached to plates installed at the tool's output ends, as shown in Fig.\ref{visionsetting}(a). Fig.\ref{visionsetting}(b) illustrates the front end. Fig.\ref{visionsetting}(c) illustrates back end. Note that since the two ends rotate identically except for their directions, we only show the measured results of the front end below.

\begin{figure}[!htbp]
\begin{center}
\includegraphics[width=0.48\textwidth]{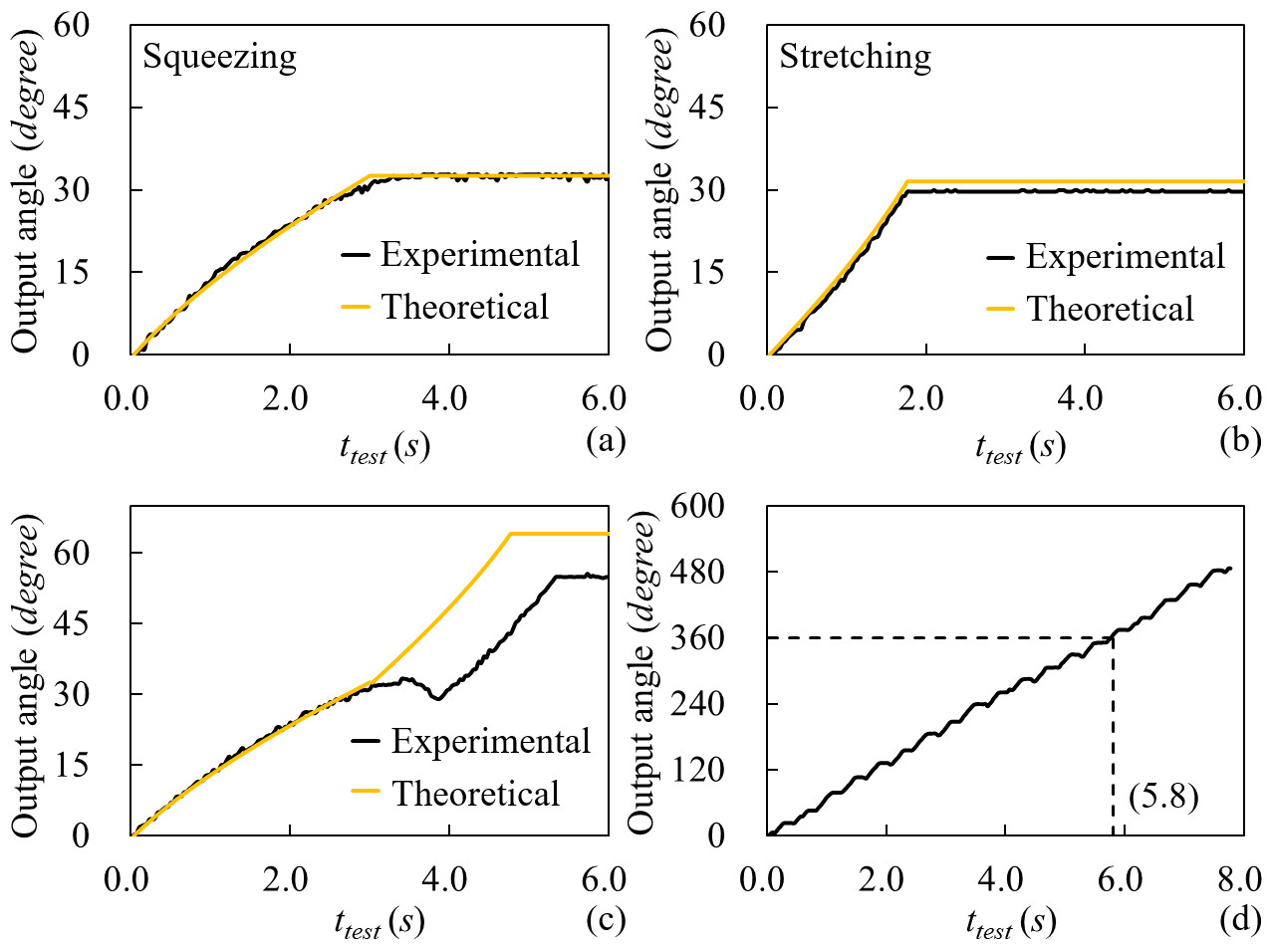}
\caption{The velocities at the tooltip. Yellow curves indicate the theoretical values. Black curves indicate measured values. (a) The changes in the rotation angle with respect to time in the squeezing phase. (b) The changes in the stretching phase. (c) The changes in a whole squeezing-stretching cycle. (d) The changes in the rotation angle during continuous squeezing and stretching. Note that the results are from the front end. The back ends rotate identically except that the direction is reversed.}
\label{exprotation}
\end{center}
\end{figure}

Fig.\ref{exprotation} shows the measured results. Like Fig.\ref{exp-output-data}, the yellow curves indicate the theoretical values, and the black curves indicate the experimental values. The four subfigures in Fig.\ref{exprotation} correspond to the results of the four cases mentioned above. Fig.\ref{exprotation}(a) and (b) are respectively the relation between the rotation angle and time in the squeezing phase and the stretching phase. For these to single-direction stroke, the results of the experiments well match the theoretical values. Fig.\ref{exprotation}(c) is the relation in a whole squeezing-stretching cycle. The curves show that the experiment results start to deviate from the theoretical values when the tooltip switches its rotation direction. Instead of a continuous smooth increase, the rotation angle keeps constant from 3.1$s$ to 3.4$s$, drops from 3.4$s$ to 3.9$s$, and resumes to increase following a pattern like Fig.\ref{exprotation}(b) after 3.9$s$. The reason for the constant rotation is that there is a short switching delay between the two phases. The reason for the drop is because of the coupling backlash and the errors in machining accuracy. The coupling backlash occurs when ratchet gears switch the directions. It blocks the motion from being transmitted to the output end.

Note that the hand speed used to measure the real angle as well as computing the theoretical values for the first three subfigures is 0\% (20$mm/s$ according to the specification of Robotiq Hand-E). This slow motion was selected since we would like to have a detailed view of the changes.

Fig\ref{exprotation}(d) further shows the results where the gripper squeezes and stretches the tool continuously. The curve is made of a sequence of smaller patterns where each of them is like the one shown in Fig.\ref{exprotation}(c). In this case, the hand speed used to measure the real angle is selected to be 100\% (150$mm/s$). The tool can output a 360$^\circ$ rotation in 5.8$s$ under this speed. 

\subsection{Real-world Applications}
Next, we develop applications to examine the manipulation policies of the tool. The video clips of these applications are available in the supplementary file of the article.

\subsubsection{Exchanging tooltips for various tasks}
\begin{figure}[!htbp]
\begin{center}
\includegraphics[width=0.48\textwidth]{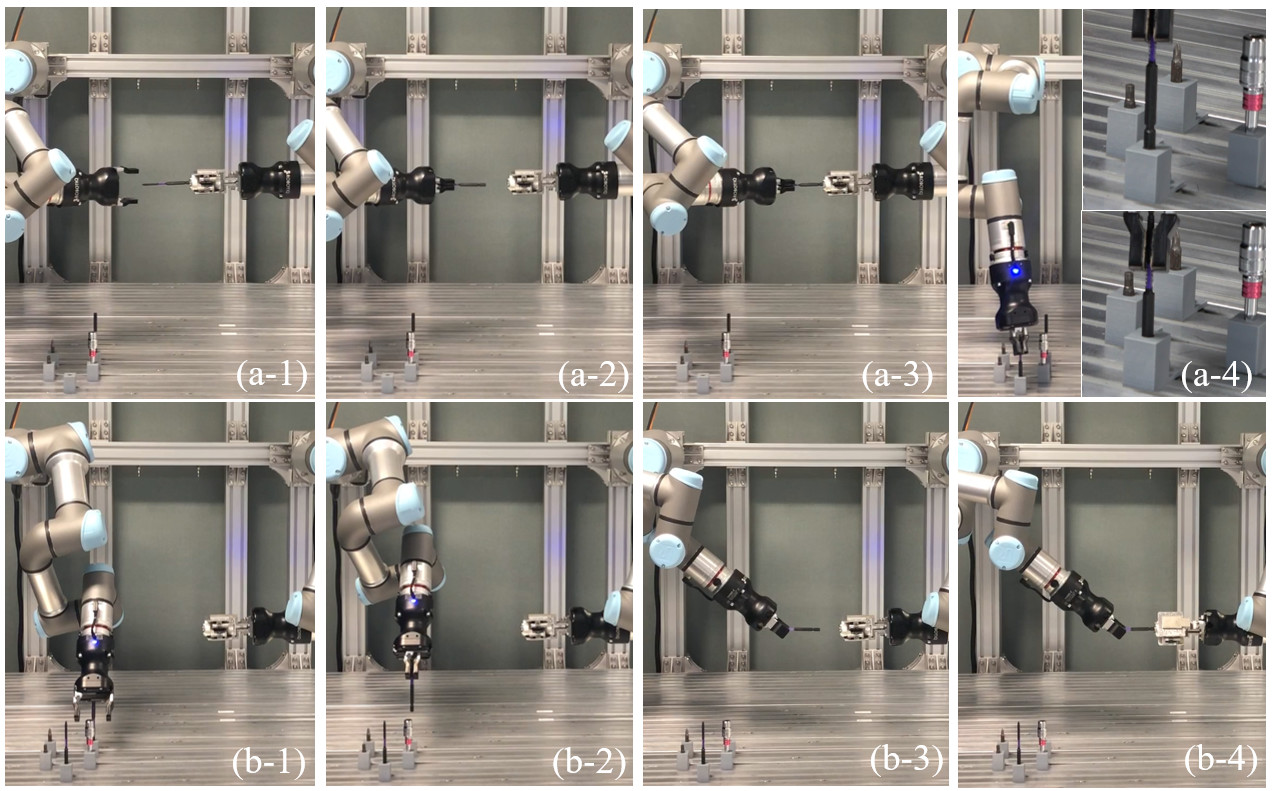}
\caption{Exchanging a \#3 hexagonal screwdriver tooltip with a \#6 one. (a) Unplugging the \#3 tooltip and returning it to the tooltip holder. (b) Picking up the \#6 tooltip and inserting it into the output socket.}
\label{Experiment-exchange}
\end{center}
\end{figure}

\begin{figure}[!htbp]
\begin{center}
\includegraphics[width=0.48\textwidth]{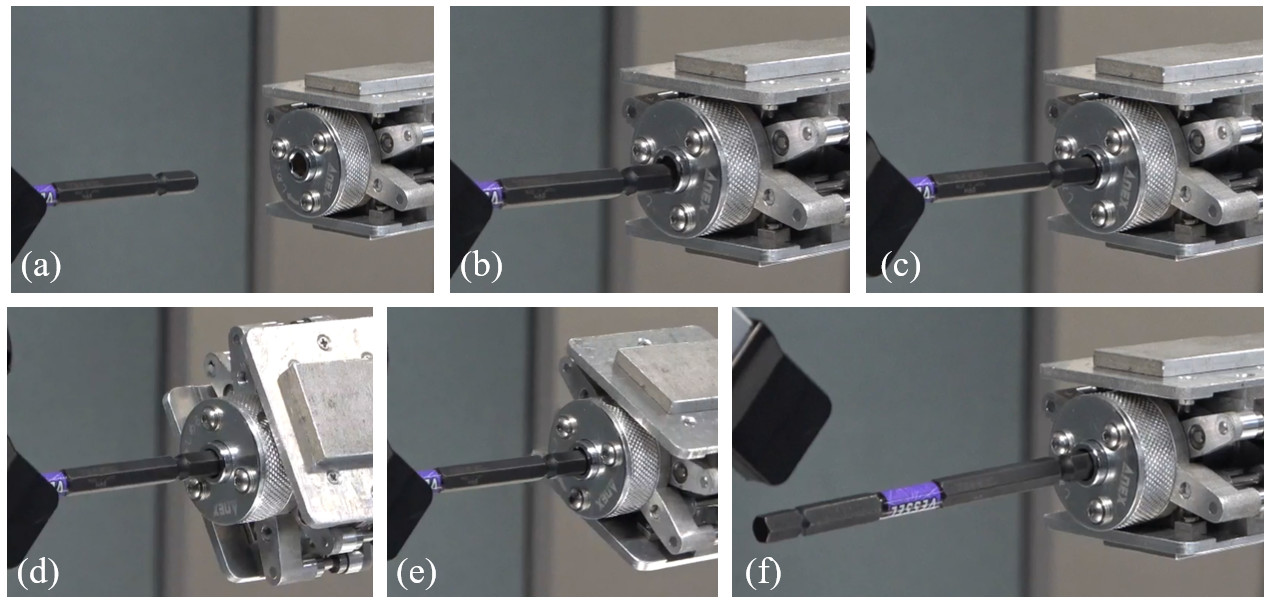}
\caption{Detailed snapshots showing the insertion of the \#6 tooltip. The linear research stars from (a) and stops at (b). The spiral research starts from (b). It stops when the hole is found at (c). (d, e) Rotation search and impedance control. (f) The insertion is done.}
\label{Experiment-insertion}
\end{center}
\end{figure}

First, we program the robot to exchange the tooltips for the tool. The goal is to remove the current \#3 hexagonal screwdriver tooltip and replace it with a \#6 one. Some snapshots showing the execution results are shown in Fig.\ref{Experiment-exchange}. With the tool held by the left hand, the right hand unplugs the \#3 tooltip and inserts a \#6 tooltip into the tool's output socket. To ensure a successful insertion, the linear research, spiral research, and rotation search and impedance control mentioned in Section IV.C are exercised. The insertion process is shown in detail in Fig.\ref{Experiment-insertion}. Here, (a) is the starting position for linear research. (b) is the starting position for spiral search. The tooltip gets in contact with the tool's output surface at the end of the linear search, and the spiral search gets started. The spiral search finishes at (c). After that, the rotation research and impedance control start, as is shown in (d, e). (f) shows the final replaced tooltip.

\begin{figure}[!htbp]
\begin{center}
\includegraphics[width=0.48\textwidth]{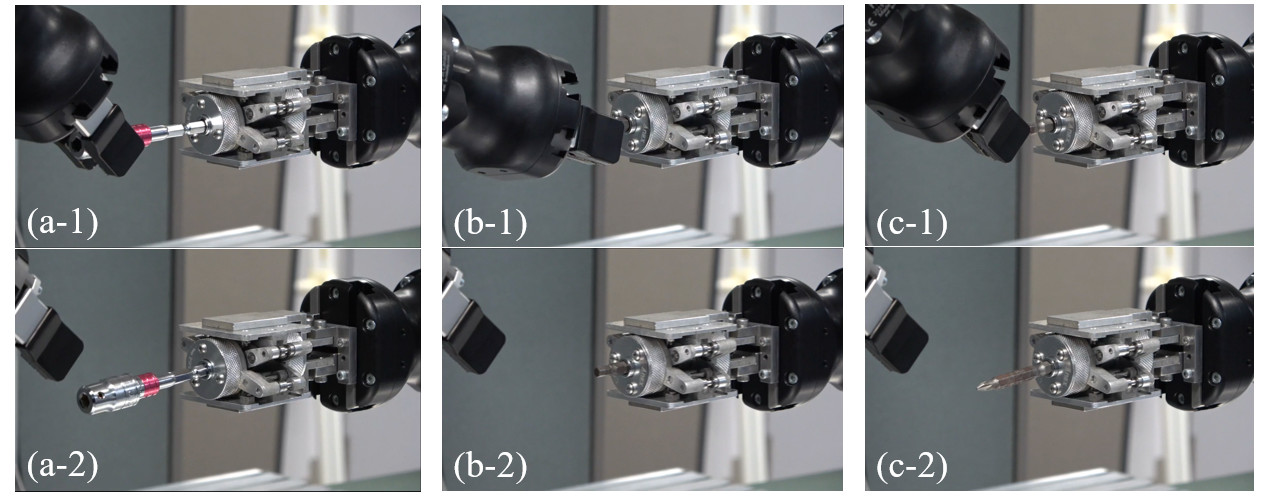}
\caption{Some other validations of the manipulation policies for exchanging tooltips. (a) Exchanging to an extra hexagonal socket. (b) Exchanging to a short hexagonal screwdriver. (c) Exchanging to a cross screwdriver.}
\label{Experiment-varioustip}
\end{center}
\end{figure}

Besides the exchanging task mentioned above, another three different tooltips, including an extra hexagonal socket, a short hexagonal screwdriver, and a cross screwdriver, are used to test the robustness and adaptability of the manipulation policies for exchanging the tooltips. The snapshots of them are shown in Fig.\ref{Experiment-varioustip}.

\subsubsection{Fastening task with automatic recognition and planning}
\begin{figure}[!htbp]
\begin{center}
\includegraphics[width=0.48\textwidth]{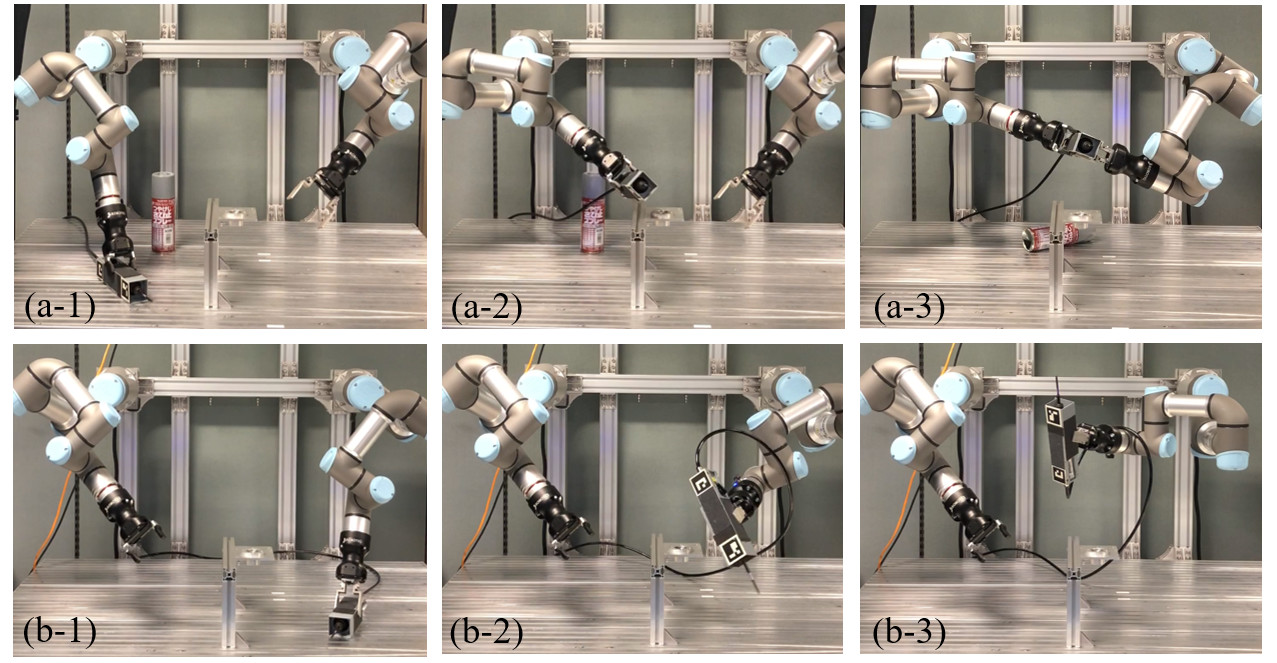}
\caption{Manipulating a vacuum driver can be difficult due to the annoying cable connection. (a) The cable (vacuum tube) knocks down an obstacle in the environment. (b) One robot arm gets entangled with the tube.}
\label{Experiment-badexample}
\end{center}
\end{figure}

Second, we program the robot to conduct fastening tasks. The goal is to verify the robustness and reliability of the tool while also demonstrating the advantages of mechanical screwing compared with other solutions.
\paragraph{Flexibility in collision-free motion planning}
We compare the flexibility of the proposed mechanical screwing tool with a conventional pneumatic screwdriver widely used in the manufacturing industry. In the case that the robot uses the pneumatic screwdriver, as shown in Fig.\ref{Experiment-badexample}, the robot has difficulty in dealing with the vacuum tube. The vacuum tube may knock down the spray bottle placed in the workspace during the manipulation, as is shown in Fig.\ref{Experiment-badexample}(a). Also, the robot may get entangled with the vacuum tube during manipulation, as is shown in Fig.\ref{Experiment-badexample}(b). Very smart modeling algorithms and motion planners must be developed to avoid these problems.

\begin{figure}[!htbp]
\begin{center}
\includegraphics[width=0.48\textwidth]{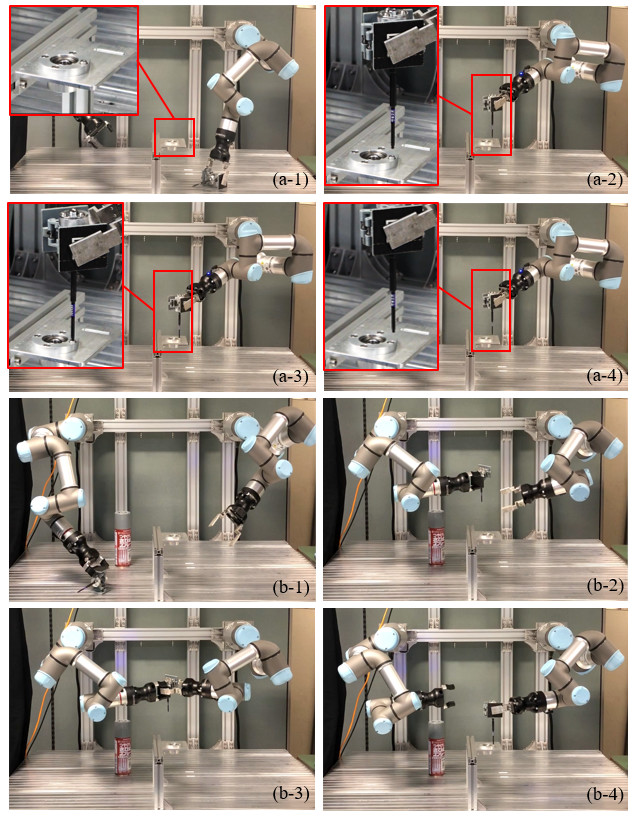}
\caption{(a) The robot uses the tool to fasten a screw. (b) Handover is applied to reorient the tool to reach a feasible tool-control grasp pose.}
\label{Experiment-motion}
\end{center}
\end{figure}

Contrarily, since our tool is mechanical, it does not have any ``tails'' like electric cables, signal wires, or vacuum tubes. There is thus no need to consider the negative influence of them. Fig.\ref{Experiment-motion} exemplifies a fastening task using the proposed tool. According to the tool's initial states, our planner found two motion sequences to finish the given task. In the first initial state shown in Fig.\ref{Experiment-motion}(a), the tool is placed on the table in a pose that the holding pads are vertical to the table. Under this state, the robot can directly pick up the tool using a tool-control pose and move the tool to the goal position to perform the fastening task. In the second initial sate shown in Fig.\ref{Experiment-motion}(b), the tool is standing on one side of the holding pad. The robot cannot grasp the tool using the tool-control pose. It thus picks up the tool using a tool-holding grasp pose and adjusts to tool-control pose by handing the tool over to the second arm. 

\begin{figure}[!htbp]
\begin{center}
\includegraphics[width=0.48\textwidth]{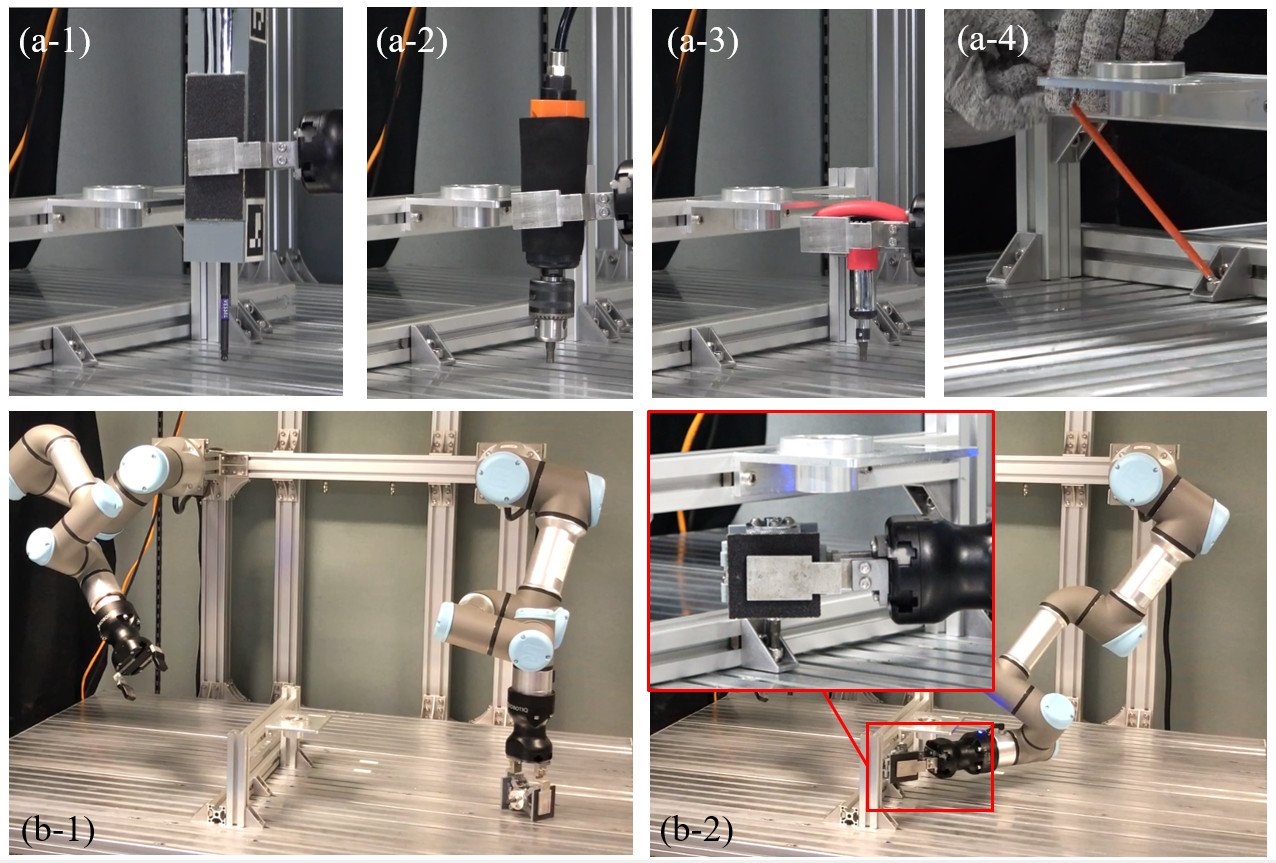}
\caption{(a) The robot cannot use commercial tools to work in a narrow space. It is even inconvenient for humans to use a commercial hexagon wrench. (b) The robot can finish the fastening task by selecting a feasible tool-control grasp pose.}
\label{Experiment-narrow}
\end{center}
\end{figure}

\paragraph{Working in narrow spaces}
We also study the ability of the tool to fasten a screw in a narrow space. Fig.\ref{Experiment-narrow} shows the scenario. The robot cannot use commercial tools to work in the narrow space shown in the scene. It is even not convenient for humans to use a commercial hexagon wrench, as shown in Fig.\ref{Experiment-narrow}(a). In contrast, the proposed tool is compact and has many tool-control grasp poses, and it can work in the narrow space to fasten the screw, as shown in Fig.\ref{Experiment-narrow}(b).

\section{CONCLUSIONS AND FUTURE WORK}
This paper presented the design, optimization, and the manipulation policies of a mechanical tool for robots with 2-finger parallel grippers. The tool can convert linear motion into rotational motion. Thus, it could be used by a robot to fasten screws. Two mC-SLE mechanisms and a double-ratchet mechanism are employed in the design. Force analysis and geometric constraints are considered to make the tool have effective transmission capabilities and a compact structure. The manipulation policies combining linear search, spiral search, and rotation search and impedance control are developed to exchange the tooltips and adjust grasp poses. Experiments show a prototype of the designed tool has a good expected performance. Robots can use the tool to exchange the tooltip and conduct various fastening tasks with visual detection and grasp and motion planning. The tool has good robustness and adaptability.

A significant problem of the tool is its efficiency. In our future work, we plan to improve the efficiency by including additional layers, i.e., reduction gear layers, to the current design.

\bibliographystyle{IEEEtran}
\bibliography{huICRA2019}

\end{document}